\documentclass[letterpaper]{article} 
\usepackage[preprint]{aaai2027}
\usepackage[hyphens]{url}  
\usepackage{graphicx} 
\usepackage{natbib}  
\usepackage{caption} 
\usepackage{multirow}
\usepackage{amssymb}
\usepackage{pifont}
\newcommand{\cmark}{\ding{51}}
\newcommand{\xmark}{\ding{55}}
\usepackage{xcolor}
\usepackage{colortbl}
\usepackage{amsmath}

\usepackage{xspace}
\newcommand{\method}{\textsc{SUV}\xspace}

\newcommand{\methodwoaccess}{\textsc{SUV} w/o Future Access\xspace}

\usepackage{booktabs}

\title{SUV: Future Scene Understanding as Video Generation for End-to-End Driving}
\author{
Yibo Yuan\textsuperscript{\rm 1}\equalcontrib,
Jiacheng Fu\textsuperscript{\rm 2}\equalcontrib,
Jiangtong Zhu\textsuperscript{\rm 3}\equalcontrib,
Yi Li\textsuperscript{\rm 3},
Jianhua Han\textsuperscript{\rm 3},\\
Meng Tian\textsuperscript{\rm 3},
Zhuohan Liu\textsuperscript{\rm 4},
Zhiwei Xiong\textsuperscript{\rm 3},
Hang Xu\textsuperscript{\rm 3},
Jianwu Fang\textsuperscript{\rm 1},
Jianru Xue\textsuperscript{\rm 1}\corresponding
}
\affiliations{
\mbox{\textsuperscript{\rm 1}Xi'an Jiaotong University}\quad
\mbox{\textsuperscript{\rm 2}University of Science and Technology of China}\\
\mbox{\textsuperscript{\rm 3}Yinwang Intelligent Technology Co., Ltd.}\quad
\mbox{\textsuperscript{\rm 4}Fudan University}\\[0.35em]
\textbf{Code:} \pdfstartlink attr{/Border [0 0 0]} user{/Subtype /Link /A << /S /URI /URI (https://github.com/ASH-2046/SUV) >>}\url{https://github.com/ASH-2046/SUV}\pdfendlink
}

\begin{document}

\maketitle

\begin{abstract}
End-to-end driving requires a coherent understanding of future scenes, yet existing methods model these scenes using task-specific heads and output formats, with limited scalability. Can video generation instead provide a shared predictor? We introduce SUV, a unified end-to-end driving framework that casts future \underline{\textbf{S}}cene \underline{\textbf{U}}nderstanding as \underline{\textbf{V}}ideo generation using a pretrained video foundation model. SUV models future appearance, semantics, relative depth, and instance-level dynamics as video streams with a shared video expert, without stream-specific visual prediction heads. Through joint video-action attention, the action expert attends to the latent representations of all future streams and generates the ego trajectory. Experiments show that SUV directly predicts all four future streams, while controlled ablations show that structured future supervision and direct future-stream access yield higher trajectory planning scores. With only a single front camera and no candidate-trajectory selection, SUV outperforms a broad set of recent state-of-the-art methods on both NAVSIM-v2 splits, achieving 91.0 EPDMS on navtest and 36.9 on navhard. On the long-tail WOD-E2E benchmark, SUV achieves a competitive RFS of 7.94.
\end{abstract}


\section{Introduction}

In end-to-end planning, a policy predicts a future ego trajectory from an observation history. This prediction requires more than understanding the observed scene. The policy must also anticipate how the scene may evolve over the planning horizon. How to represent this future scene understanding and make it directly usable for trajectory generation remains a central challenge.

Existing end-to-end paradigms address this problem from different directions. Multi-task frameworks such as UniAD~\cite{uniad}, VAD~\cite{vad}, and SparseDrive~\cite{sparsedrive} jointly optimize perception, mapping, motion prediction, and planning. Each task usually requires its own queries, predictor, decoder, and objective, with limited module scalability. Vision-language-action models (VLAs) instead use a common language interface to express different driving outputs~\cite{drivevlm,omnidrive,senna,emma}. This design makes task expansion easier, but discrete language tokens struggle to capture dense spatial structure and continuous scene motion. Some VLAs augment the policy with future-image prediction or latent world states~\cite{drivevlaw0,driveworldvla}. World-action models (WAMs) directly couple future prediction with action generation~\cite{hou2026worldmodel,motus,tesseract,fastwam}. Most methods only involve pixel-level RGB video or latent vision-state prediction~\cite{drivinggpt,doe1,law,drivelaw,metis,drivefuture}. Yet future scenes contain not only RGB appearance but also road semantics, geometry, and instance-level dynamics. To expose these properties, existing methods usually add task-specific branches, readouts, or auxiliary objectives~\cite{xwam,wam4d,drivedreamerpolicy,eponav2}, falling again into the multi-task paradigms.
\begin{figure}[t]
\centering
\includegraphics[width=\linewidth]{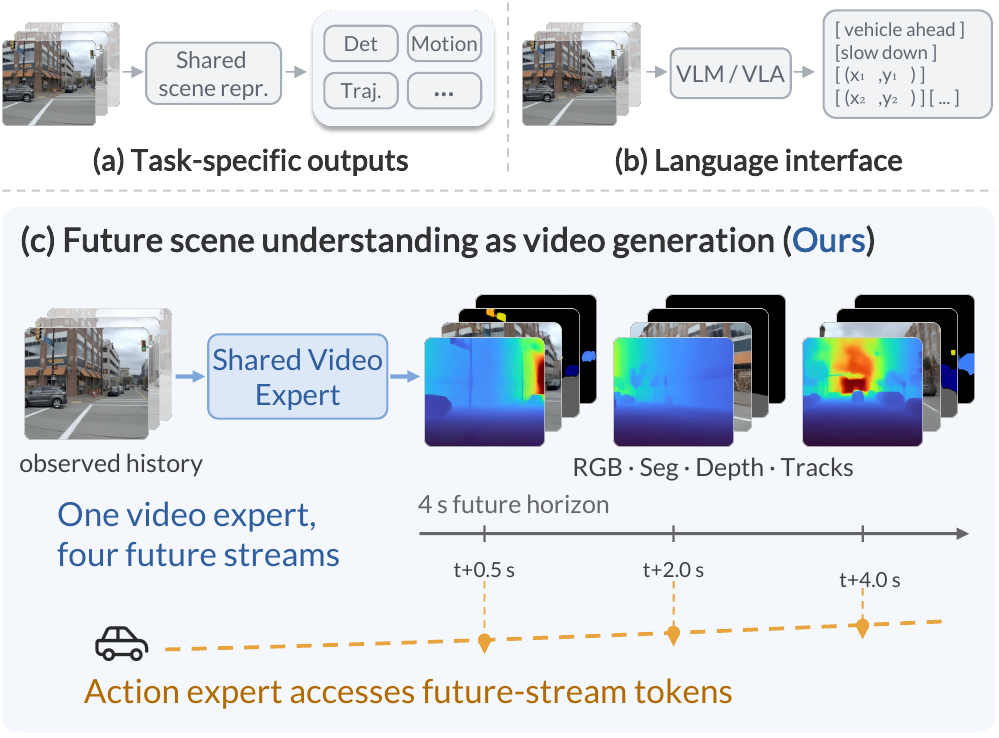}
\caption{Future scene understanding for end-to-end driving. (a) Multi-task
methods share a scene representation but retain task-specific outputs.
(b) Language-based methods serialize scene information and trajectories as
tokens. (c) \method uses one shared video expert to generate four future
streams and lets the action expert attend to their latent tokens for
planning.}
\label{fig:teaser}
\end{figure}

Large video generators such as Wan~\cite{wan2025} and Cosmos~\cite{nvidia2025cosmosworldfoundationmodel} learn spatial structure and temporal dynamics from large-scale video data. Recent work builds on these capabilities by adapting pretrained video generators to depth estimation, segmentation, pose estimation, and tracking~\cite{wang2026videogeneration}. Other studies instead use image generation as a shared interface for visual understanding tasks with heterogeneous outputs~\cite{gabeur2026imagegenerators,han2026visionunified}. These studies leave open whether a single pretrained video generator can jointly model diverse aspects of future scene understanding for end-to-end driving.

We introduce \method, an end-to-end driving framework that formulates future \textbf{\underline{S}}cene \textbf{\underline{U}}nderstanding as \textbf{\underline{V}}ideo generation (Fig.~\ref{fig:teaser}(c)). To the best of our knowledge, \method is the first end-to-end driving framework to generate RGB, segmentation, relative-depth, and instance-track futures as native video streams with a shared video expert. We initialize the video expert from the pretrained Wan2.2-5B generator~\cite{wan2025} and post-train it in the generator's native latent video space. The expert jointly generates all four streams over a 4-s horizon without stream-specific visual prediction heads. The four streams represent future appearance, semantic layout, scene geometry, and instance-level dynamics, respectively. During joint denoising, the action expert attends to future-stream latents and generates the ego trajectory.

Using only a single front camera, \method outperforms a broad set of recent state-of-the-art methods on both NAVSIM-v2~\cite{navsimv2} splits, achieving 91.0 EPDMS on navtest and 36.9 on navhard. On the long-tail WOD-E2E benchmark~\cite{wode2e}, \method achieves a competitive RFS of 7.94. Our \textbf{contributions} are summarized as follows.

\begin{itemize}
    \item We formulate future scene understanding as video generation and post-train a single video foundation model to generate RGB, segmentation, relative-depth, and instance-track futures without stream-specific heads.
    
    \item We connect these future representations to trajectory planning through masked joint video-action attention, allowing a separate action expert to attend to all future-stream latents while generating the ego trajectory.
    
    \item We evaluate \method on standard and long-tail planning benchmarks and use controlled ablations to isolate how structured future supervision and direct future-stream access affect planning.
\end{itemize}

\section{Related Work}

\paragraph{Task interfaces for autonomous driving.} Multi-task end-to-end frameworks organize perception, prediction, mapping, and planning around shared bird's-eye-view (BEV), vectorized, or sparse scene representations~\cite{uniad,vad,sparsedrive}. Their shared architectures enable cross-task interaction, but heterogeneous outputs remain tied to task-specific predictors and decoding formats. Vision-language models (VLMs) and vision-language-action models (VLAs) instead express driving outputs in the language space of pretrained multimodal models~\cite{drivevlm,omnidrive,senna,automot}. EMMA serializes trajectories, 3D objects, and road graphs as text~\cite{emma}, while DriveVLA-W0 and DriveWorld-VLA augment VLA planners with visual or latent world modeling~\cite{drivevlaw0,driveworldvla}. \method differs by generating heterogeneous future scene signals as native video streams with one pretrained video expert.

\paragraph{Generative foundation models for visual understanding.} Large-scale video generators such as Wan and Cosmos learn spatial structure and temporal dynamics through generative pretraining~\cite{wan2025,nvidia2025cosmosworldfoundationmodel}. Independent work in general vision reveals two uses of generative models beyond synthesis. GenCeption transfers video-generative priors to depth, segmentation, pose, and tracking~\cite{wang2026videogeneration}. Vision Banana and SenseNova-Vision instead use generative models to produce heterogeneous visual outputs~\cite{gabeur2026imagegenerators,han2026visionunified}. These studies show that generative pretraining can support both transferable perception and shared modeling across visual tasks. \method extends this direction to future scene understanding in end-to-end driving, where the generated streams must also inform planning.

\paragraph{World-action models.} Driving world models first treat future RGB video as a controllable representation~\cite{gaia1,drivedreamer,drivewm,vista}, while WAMs couple prediction and action generation~\cite{hou2026worldmodel,motus,tesseract}. ImagiDrive feeds trajectory-conditioned future frames back to refine trajectories~\cite{imagidrive}, whereas LMGenDrive co-trains multi-view video and control but omits diffusion generation at inference~\cite{lmgendrive}. Fast-WAM and Metis use video only as training supervision~\cite{fastwam,metis}. Existing methods also differ in the representations they predict for future scenes. UniFuture forecasts RGB and depth without planning~\cite{unifuture}, whereas ExploreVLA uses RGB and depth futures for dense supervision and prediction uncertainty for exploration~\cite{explorevla}. EponaV2, GeoSem-WAM, X-WAM, and WAM4D add depth or semantics through dedicated heads or readouts~\cite{eponav2,geosemwam,xwam,wam4d}. Feature-space alternatives include SeerDrive's iterative future-BEV planning and Drive-JEPA's non-generative predictive embeddings~\cite{seerdrive,drivejepa}. \method combines native video generation of RGB, segmentation, relative-depth, and instance-track futures in one pretrained expert without stream-specific visual heads, while letting the action expert read all streams during joint denoising.
\section{Method}

\begin{figure*}[t]
\centering
\includegraphics[width=0.85\linewidth]{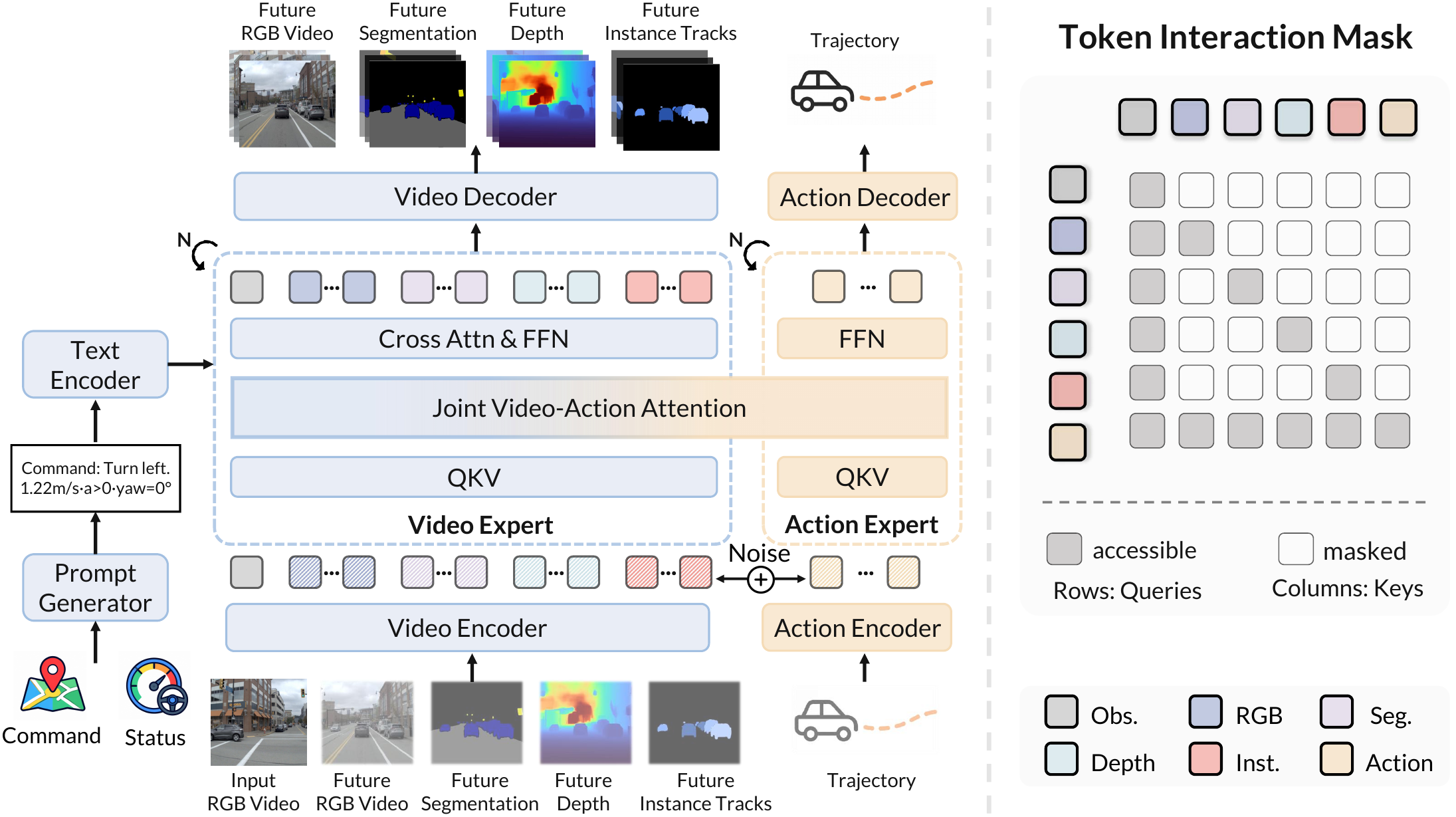}
\caption{Overview of \method. Left: the shared video expert generates four
future streams, while the action expert denoises the trajectory using their
latent tokens. Right: the token-interaction mask preserves
observation-to-stream pathways, blocks cross-stream interaction and
action-to-future feedback, and lets action queries read every token group. The
label Obs.\ denotes observation tokens encoded from the input RGB video.}
\label{fig:method}
\end{figure*}

As shown in Fig.~\ref{fig:method}, \method models future scene understanding and trajectory planning as a joint video-action generation task. A shared video expert generates RGB, segmentation, relative-depth, and instance-track futures that represent future appearance, semantic layout, scene geometry, and instance-level dynamics. A separate action expert denoises the trajectory tokens while attending to the latent tokens of all four future streams through masked joint video--action attention. At inference, \method jointly denoises the future streams and ego trajectory in latent space. The action expert therefore uses future-stream latents rather than decoded videos.
\subsection{Problem Formulation}

At time \(t\), the model receives a \(K\)-frame front-camera RGB history \(\mathbf{o}_t=\{o_{t-K+1},\ldots,o_t\}\), a navigation command \(\mathbf{g}_t\), and the current ego state \(\mathbf{s}_t\). We denote the driving context by \(\mathbf{c}_t=(\mathbf{g}_t,\mathbf{s}_t)\). The model predicts an \(H\)-step ego-frame trajectory \(\mathbf{a}_t=\{(x_{t+h},y_{t+h},\psi_{t+h})\}_{h=1}^{H}\), where \((x,y)\) and \(\psi\) denote planar position and heading.

Let \(\mathcal{M}=\{\mathrm{rgb},\mathrm{seg},\mathrm{depth},\mathrm{track}\}\) denote the four future-scene stream types. For each \(m\in\mathcal{M}\), \(\mathbf{v}_t^m=\{v_{t+j}^m\}_{j=1}^{T}\) denotes a \(T\)-frame future video, and \(\mathbf{V}_t=\{\mathbf{v}_t^m\}_{m\in\mathcal{M}}\) collects all four streams. With parameters \(\theta\), \method models their joint conditional distribution with the ego trajectory as
\begin{equation}
p_{\theta}(\mathbf{V}_t,\mathbf{a}_t\mid\mathbf{o}_t,\mathbf{c}_t).
\label{eq:joint_prediction}
\end{equation}
The four visual streams describe the same future interval using the same camera view, image grid, and frame timestamps. \method jointly denoises these streams and the trajectory, while directed attention allows the action tokens to read the evolving future-scene tokens.

\subsection{Future Scene Understanding as Video Generation}

\paragraph{A common video interface.}
We encode each future target modality as a three-channel video with the same camera-aligned image grid, prediction horizon, and timestamps as the future RGB stream. This shared representation preserves the generator's native video interface while extending its pretrained generative capabilities beyond RGB. We use recorded future camera frames for the RGB stream and render semantic labels, relative depth, and instance tracks as separate three-channel video streams.

\paragraph{Structured target construction.} A frozen Segment Anything Model 3 (SAM~3)~\cite{carion2025sam3segmentconcepts} provides semantic masks and temporally associated instance masks. We render semantic classes with a fixed palette. Each instance track receives a deterministic color from a class-specific palette and retains it throughout the future clip. A frozen Depth Anything 3 (DA3)~\cite{depthanything3} provides relative depth, which we robustly normalize within each clip, quantize, and render with a fixed color map. We decode generated structured videos only for future-scene evaluation. The action expert instead reads their evolving latent tokens during joint denoising. Exact normalization bounds, quantization rules, color palettes, and decoding procedures are provided in the supplementary material.

\paragraph{Shared latent video space.} A frozen video VAE with encoder \(\mathcal{E}_{\mathrm{vae}}\) encodes the observation history and all four future streams:
\begin{equation}
\mathbf{z}_t^{o}=\mathcal{E}_{\mathrm{vae}}(\mathbf{o}_t),
\qquad
\mathbf{z}_t^{m}=\mathcal{E}_{\mathrm{vae}}(\mathbf{v}_t^m),
\quad m\in\mathcal{M}.
\label{eq:video_encoding}
\end{equation}
The four future streams follow a shared observation prefix. All streams use the same latent layout and temporal coordinates. A frozen text encoder maps stream-specific prompts to embeddings that identify the prediction target of each stream. A shared video expert predicts the latent representations of all four streams under the same flow-matching formulation, and the shared VAE decoder decodes each representation into a video. This design requires no stream-specific visual prediction heads.

\subsection{Joint Future-Action Generation}
\definecolor{groupgray}{gray}{0.93}
\newcommand{\navsimvtesttable}{%
\centering
\small
\setlength{\tabcolsep}{5.7pt}
\begin{tabular}{l|c|cccc|ccccc|c}
\toprule
Method & Sensors & NC$\uparrow$ & DAC$\uparrow$ & DDC$\uparrow$ & TLC$\uparrow$ & EP$\uparrow$ & TTC$\uparrow$ & LK$\uparrow$ & HC$\uparrow$ & EC$\uparrow$ & \textbf{EPDMS}$\uparrow$ \\
\midrule
Human Agent & -- & 100 & 100 & 99.8 & 100 & 87.4 & 100 & 100 & 98.1 & 90.1 & 94.5 \\
\midrule
\rowcolor{groupgray}
\multicolumn{12}{c}{\textit{Traditional E2E policies}} \\
DiffusionDriveV2~\cite{diffusiondrivev2} & 3$\times$C+L & 97.7 & 96.6 & 99.2 & \underline{99.8} & 88.9 & 97.2 & 96.0 & 97.8 & \textbf{91.0} & 87.5 \\
DriveSuprim~\cite{drivesuprim} & 3$\times$C & 98.4 & \underline{98.6} & \underline{99.6} & \underline{99.8} & \underline{90.5} & 97.8 & 97.0 & \underline{98.3} & 78.6 & 87.1 \\
SparseDriveV2~\cite{sparsedrivev2} & 3$\times$C & 98.1 & 98.1 & \underline{99.6} & \underline{99.8} & \textbf{91.1} & 97.3 & 96.9 & 98.2 & 78.4 & 90.1 \\
\midrule
\rowcolor{groupgray}
\multicolumn{12}{c}{\textit{VLA-based policies}} \\
DriveVLA-W0~\cite{drivevlaw0} & 1$\times$C & 98.4 & 95.2 & 99.4 & \textbf{99.9} & 86.6 & 97.9 & 97.8 & \underline{98.3} & 82.7 & 86.9 \\
DriveWorld-VLA~\cite{driveworldvla} & 3$\times$C & 98.6 & \textbf{99.1} & \underline{99.6} & \underline{99.8} & 87.4 & 97.9 & 97.0 & 97.8 & 78.6 & 86.8 \\
SGDrive~\cite{sgdrive} & 1$\times$C & 98.6 & 94.3 & 99.5 & \underline{99.8} & 86.0 & 97.9 & 96.1 & \underline{98.3} & 85.9 & 86.2 \\
DriveFine~\cite{drivefine} & 1$\times$C & \underline{98.7} & 97.3 & 99.5 & \underline{99.8} & 88.7 & 97.8 & 97.7 & \textbf{98.4} & 83.8 & 89.7 \\
\midrule
\rowcolor{groupgray}
\multicolumn{12}{c}{\textit{WAM-based policies}} \\
EponaV2~\cite{eponav2} & 1$\times$C & 98.5 & 97.4 & 99.5 & \textbf{99.9} & 87.9 & \underline{98.1} & 97.7 & 98.2 & 77.4 & 88.9 \\
Metis~\cite{metis} & 1$\times$C & 98.4 & 97.2 & \underline{99.6} & \underline{99.8} & 87.8 & 97.7 & 97.8 & \textbf{98.4} & 88.0 & 89.5 \\
Metis~(Top 6)~\cite{metis} & 1$\times$C & 98.5 & 97.5 & \underline{99.6} & \underline{99.8} & 87.9 & 97.8 & \underline{98.0} & \textbf{98.4} & \underline{90.0} & \underline{90.3} \\
\midrule
\textbf{\method}~(Ours) & 1$\times$C & \textbf{99.1} & 97.8 & \textbf{99.7} & \underline{99.8} & 87.8 & \textbf{98.7} & \textbf{98.1} & \textbf{98.4} & 88.3 & \textbf{91.0} \\
\bottomrule
\end{tabular}
\caption{End-to-end trajectory planning on NAVSIM-v2 navtest. Results use
the corrected official EPDMS implementation. C and L denote camera and
   LiDAR. Bold and underlined values mark the best and second-best learned
   results.}\label{tab:navsim_v2}
}

\newcommand{\navsimvhardtable}{%
\centering
\small
\setlength{\tabcolsep}{5.5pt}
\begin{tabular}{l|c|cccc|ccccc|c|c}
\toprule
Method & Stage & NC$\uparrow$ & DAC$\uparrow$ & DDC$\uparrow$ & TLC$\uparrow$ & EP$\uparrow$ & TTC$\uparrow$ & LK$\uparrow$ & HC$\uparrow$ & EC$\uparrow$ & S.$\uparrow$ & \textbf{EPDMS}$\uparrow$ \\
\midrule
\rowcolor{groupgray}
\multicolumn{13}{c}{\textit{Traditional E2E policies}} \\
\multirow{2}{*}{LTF~\cite{transfuser}}
& S1 & 96.2 & 79.6 & 99.1 & 99.6 & 84.1 & 95.1 & 94.2 & 97.6 & \underline{79.1} & -- & \\
& S2 & 77.8 & 70.2 & 84.3 & 98.1 & 85.1 & \textbf{85.1} & 45.4 & 95.7 & \textbf{76.0} & -- & \multirow{-2}{*}{25.1} \\
\midrule
\multirow{2}{*}{GuideFlow~\cite{guideflow}}
& S1 & 96.6 & 80.5 & 96.3 & 99.3 & 82.3 & 94.9 & 91.5 & \underline{97.7} & 67.8 & -- & \\
& S2 & \textbf{87.3} & \underline{76.7} & \textbf{88.8} & \textbf{99.2} & 84.3 & \textbf{85.1} & \underline{49.7} & 93.1 & 44.5 & -- & \multirow{-2}{*}{27.1} \\
\midrule
\rowcolor{groupgray}
\multicolumn{13}{c}{\textit{VLA-based policies}} \\
\multirow{2}{*}{DriveVLA-W0~\cite{drivevlaw0}}
& S1 & 96.8 & 83.3 & 99.0 & 99.6 & \textbf{84.6} & 95.3 & 96.4 & 97.6 & 78.2 & -- & \\
& S2 & 76.8 & 64.3 & 79.9 & 98.3 & \textbf{89.2} & 75.0 & 46.8 & 95.8 & 53.1 & -- & \multirow{-2}{*}{24.4} \\
\midrule
\multirow{2}{*}{ReCogDrive~\cite{recogdrive}}
& S1 & 96.4 & 78.9 & 98.7 & \underline{99.8} & 82.6 & \underline{95.6} & 94.4 & 97.6 & 74.2 & 67.7 & \\
& S2 & 80.2 & 65.0 & 82.4 & 98.7 & 85.2 & 76.9 & 43.8 & \textbf{96.6} & 71.8 & 37.6 & \multirow{-2}{*}{25.7} \\
\midrule
\multirow{2}{*}{SGDrive~\cite{sgdrive}}
& S1 & 95.8 & 87.6 & 97.8 & \underline{99.8} & 84.4 & 94.7 & 92.9 & \textbf{97.8} & 28.9 & 71.1 & \\
& S2 & 79.4 & 65.4 & 79.1 & \underline{98.9} & \underline{88.9} & 75.3 & 42.7 & \underline{96.4} & 29.6 & 35.2 & \multirow{-2}{*}{25.5} \\
\midrule
\rowcolor{groupgray}
\multicolumn{13}{c}{\textit{WAM-based policies}} \\
\multirow{2}{*}{EponaV2~\cite{eponav2}}
& S1 & \textbf{97.3} & \underline{90.7} & \textbf{99.4} & \textbf{100} & 83.3 & \textbf{97.3} & \underline{97.3} & 97.6 & 60.9 & -- & \\
& S2 & \underline{83.6} & \textbf{78.0} & \underline{88.0} & \underline{98.9} & 86.0 & \underline{80.3} & \textbf{50.1} & 96.1 & 52.0 & -- & \multirow{-2}{*}{\underline{36.1}} \\
\midrule
\multirow{2}{*}{Metis~\cite{metis}}
& S1 & 96.6 & 87.8 & 99.0 & 99.3 & \underline{84.5} & \underline{95.6} & \textbf{97.8} & \textbf{97.8} & 77.8 & \underline{75.8} & \\
& S2 & 79.6 & 73.3 & 84.9 & 97.8 & 85.8 & 76.6 & 47.7 & 95.4 & \underline{75.3} & \underline{41.7} & \multirow{-2}{*}{32.2} \\
\midrule
\multirow{2}{*}{\textbf{\method}~(Ours)}
& S1 & \underline{96.9} & \textbf{94.2} & \underline{99.3} & 99.6 & 84.1 & \underline{95.6} & 96.7 & \textbf{97.8} & \textbf{79.6} & \textbf{82.3} & \\
& S2 & 82.7 & 74.2 & 85.9 & 98.4 & 86.0 & 78.9 & 47.2 & 95.9 & 69.1 & \textbf{43.9} & \multirow{-2}{*}{\textbf{36.9}} \\
\bottomrule
\end{tabular}
\caption{End-to-end trajectory planning on NAVSIM-v2 navhard. Stage-1,
Stage-2, and combined results follow the official protocol and are ranked
separately. S. is the per-stage score. Bold and underlined values
mark the best and second-best results.}\label{tab:navsim_v2_navhard}
}

\newcommand{\navsimmainplanningtables}{%
\begin{table*}[!t]
{\navsimvtesttable}
\par\medskip
{\navsimvhardtable}
\end{table*}
}

\paragraph{Video and action experts.}
\method combines a Wan2.2-5B-initialized video expert~\cite{wan2025} shared across four future streams with a separate action expert in a Mixture-of-Transformers architecture~\cite{liang2025mixtureoftransformers}. The video expert processes a clean observation prefix and four future token groups, while the action expert represents the normalized trajectory \(\widetilde{\mathbf{a}}=\mathcal{N}_{a}(\mathbf{a})\) as waypoint tokens, where \(\mathcal{N}_{a}\) denotes trajectory normalization. Both experts cross-attend to the same common prompt, which contains the navigation command and current ego state. Each future stream additionally cross-attends to a modality-specific embedding that identifies its prediction target as RGB, segmentation, relative depth, or instance tracks. Masked joint self-attention then controls information flow among all token groups.

\paragraph{Directed future-to-action attention.}
As illustrated by the token interaction mask in the right panel of Fig.~\ref{fig:method}, we impose asymmetric information flow among token groups. Let \(g(i)\in\{\mathrm{obs},\mathrm{act}\}\cup\mathcal{M}\) denote the group of token \(i\), and let \(\mathcal{A}(g)\) specify the visible key groups:
\begin{equation}
\mathcal{A}(g)=
\begin{cases}
\{\mathrm{obs}\}, & g=\mathrm{obs},\\
\{\mathrm{obs},m\}, & g=m,\;m\in\mathcal{M},\\
\{\mathrm{obs},\mathrm{act}\}\cup\mathcal{M}, & g=\mathrm{act}.
\end{cases}
\label{eq:directed_attention}
\end{equation}
For query token \(i\) and key token \(j\), we set \(M_{ij}=0\) when \(g(j)\in\mathcal{A}(g(i))\) and \(M_{ij}=-\infty\) otherwise. Thus, observation queries read only the clean prefix, each future stream reads the prefix and itself, and action queries read all groups. Applied in every Transformer block and denoising step, this mask exposes future-scene latents to the action expert before video decoding, while blocking cross-stream interaction and action-to-future feedback.

\paragraph{Preserving pretrained attention structure.} Cross-stream attention could coordinate future streams but would introduce interactions absent from video pretraining. We therefore retain the pretrained observation-to-future attention pathway for each stream. The shared observation prefix, spatial grid, timestamps, and video expert provide all streams with a common scene context and spatiotemporal prior without direct token exchange. Figure~\ref{fig:qualitative_results} shows consistent road boundaries and vehicle locations across the streams.

\subsection{Training and Inference}

\paragraph{Multi-stream flow matching.}
Let \(\mathcal{R}=\mathcal{M}\cup\{\mathrm{act}\}\), with clean targets \(\mathbf{y}^{m}=\mathbf{z}^{m}\) and \(\mathbf{y}^{\mathrm{act}}=\widetilde{\mathbf{a}}\). The visual groups share \(\lambda_{\mathrm{vid}}\), while \(\lambda_{\mathrm{act}}\) is sampled independently. Writing \(\lambda_r\) for the corresponding time, we draw \(\boldsymbol{\epsilon}^{r}\sim\mathcal{N}(\mathbf{0},\mathbf{I})\), where \(\mathbf{I}\) is the identity covariance, and form
\begin{equation}
\mathbf{y}_{\lambda_r}^{r}
=(1-\lambda_r)\mathbf{y}^{r}
+\lambda_r\boldsymbol{\epsilon}^{r},
\qquad
\mathbf{u}^{r}
=\boldsymbol{\epsilon}^{r}-\mathbf{y}^{r}.
\end{equation}
Conditioned on \(\lambda_r\), one masked forward pass predicts \(\hat{\mathbf{u}}_{\theta}^{r}\) for all groups. We optimize
\begin{equation}
\begin{aligned}
\mathcal{L}_{\mathrm{FM}}^{r}
&=w(\lambda_r)\operatorname{MSE}
\left(\hat{\mathbf{u}}_{\theta}^{r},\mathbf{u}^{r}\right),\\
\mathcal{L}
&=\frac{1}{4}\sum_{m\in\mathcal{M}}\mathcal{L}_{\mathrm{FM}}^{m}
+\mathcal{L}_{\mathrm{FM}}^{\mathrm{act}},
\end{aligned}
\end{equation}
Here \(\operatorname{MSE}\) denotes mean-squared error and \(w\) is the normalized scheduler weight. We jointly train both experts and the ego-state projection while keeping the video VAE and text encoder frozen.

\paragraph{Joint inference.}
During inference, we synchronize the video and action flow times on a common shifted grid from \(1\) to \(0\). The observation prefix remains fixed, while all five prediction groups start from Gaussian noise. At each solver step, \method predicts the velocity for every group in a masked forward pass while the action expert reads the current future-scene latents. The solver then updates all groups using these predictions. After the final step, \(\mathcal{N}_{a}^{-1}\) restores the ego-frame trajectory. Planning remains in latent space. The VAE decoder is used only for evaluation or visualization. Trajectory normalization, scheduler, and Euler details are provided in the supplementary material.
\section{Experiments}
We investigate the following research questions:
\textbf{1)} How does \method perform across standard and long-tail planning benchmarks?
\textbf{2)} Can one video expert jointly predict all four future streams without compromising RGB quality, and how important is video-generative pretraining?
\textbf{3)} How do structured future supervision and direct future-stream access affect planning?
\textbf{4)} How does \method balance planning performance and inference efficiency?

\navsimmainplanningtables

\subsection{Experimental Setup}

\paragraph{Implementation details.}
On NAVSIM, the model takes four front-camera frames sampled at 2\,Hz over a 2-s observation window at \(640\times384\) resolution. It predicts eight future frames and eight trajectory waypoints over the next 4\,s. The action expert has a hidden dimension of 1024 and contains approximately 1B parameters. The full model contains approximately 6B parameters. We train \method for 60 epochs on eight NVIDIA H200 GPUs using AdamW with a learning rate of \(1\times10^{-4}\) and weight decay of \(0.01\). We use a cosine learning-rate schedule, bfloat16 precision, gradient clipping at \(1.0\), and random seed 42. Unless stated otherwise, inference uses 10 solver steps.

\paragraph{Planning evaluation.}
We train \method on NAVSIM navtrain and evaluate it on the NAVSIM-v2 navtest and navhard splits using the corrected official EPDMS implementation~\cite{navsim,navsimv2}. Navtest uses a one-stage protocol, whereas navhard combines 450 original observations in Stage~1 with 5{,}462 pre-generated 3DGS observations in Stage~2. We also train and evaluate \method on the long-tail WOD-E2E benchmark following its official protocol~\cite{wode2e}. WOD-E2E evaluates 5-s planning using RFS and RFS-GT ADE at 3 and 5\,s.

\paragraph{Future-scene evaluation.}
We evaluate all four future streams on 12{,}146 samples from NAVSIM-v2 navtest. We assess RGB, segmentation, and relative depth at eight horizons from 0.5 to 4.0\,s, and instance tracking at seven horizons from 1.0 to 4.0\,s. For each metric, we report the unweighted mean over its evaluation horizons. We use PSNR and SSIM for RGB, mIoU for segmentation, \(\delta_1\) and AbsRel for relative depth, and AssA@50 for instance tracking. SAM~3 provides the segmentation and tracking references, whereas DA3 provides the relative-depth references. We evaluate segmentation and tracking within a predefined 50-m region and relative depth over the full image. The segmentation, tracking, and depth metrics measure agreement with frozen teachers rather than accuracy against independent ground truth. Generate-then-Perceive applies the frozen teachers to its predicted future RGB sequence, whereas the same teachers derive the evaluation references from the recorded future RGB frames. The baseline uses the same horizons, regions, references, and metrics as native generation, providing a task-aligned comparison between native structured generation and post hoc recovery. The supplementary material provides detailed target-construction and decoding procedures.

\subsection{Planning Performance Across Benchmarks}

Using only a single front camera, \method outperforms a broad set of recent state-of-the-art methods on both NAVSIM-v2 splits. On NAVSIM-v2 navtest, \method achieves 91.0 EPDMS (Table~\ref{tab:navsim_v2}). Metis with top-6 trajectory selection achieves 90.3 EPDMS, 0.7 points lower. At the component level, \method records the highest NC, DDC, TTC, and LK scores and ties for the highest HC score. Other methods record higher EP or EC scores. On navhard, \method achieves 36.9 EPDMS (Table~\ref{tab:navsim_v2_navhard}). EponaV2 achieves 36.1 EPDMS, 0.8 points lower.

On the long-tail WOD-E2E benchmark, \method achieves a competitive RFS of 7.94 (Table~\ref{tab:wod_e2e}). It records RFS-GT ADE values of 1.24 and 2.90 at 3 and 5\,s, respectively. IRL-VLA records lower RFS-GT ADE values of 1.22 and 2.82 at the same horizons. RFS rewards trajectories judged appropriate for a scene, whereas RFS-GT ADE measures geometric deviation from the recorded expert trajectory. The two metrics need not rank methods identically because a scene can admit multiple reasonable trajectories.

\begin{table}[!t]
\centering
\small
\setlength{\tabcolsep}{5pt}
\renewcommand{\arraystretch}{1.08}
\def\adepair#1#2{\makebox[2.25em][r]{#1}\,/\,\makebox[2.55em][r]{#2}}
\begin{tabular}{@{}l@{\extracolsep{\fill}}cc@{}}
\toprule
Method & ADE 3/5$\downarrow$ & \textbf{RFS}$\uparrow$ \\
\midrule
AutoVLA~\cite{autovla} & \adepair{1.35}{2.96} & 7.56 \\
HMVLM~\cite{hmvlm} & \adepair{1.33}{3.07} & 7.74 \\
Fast-dDrive~\cite{fastddrive} & \adepair{1.25}{2.91} & 7.82 \\
IRL-VLA~\cite{irlvla} & \adepair{\textbf{1.22}}{\textbf{2.82}} & 7.89 \\
Poutine-Base~\cite{poutine} & \adepair{1.27}{2.94} & \underline{7.91} \\
\midrule
\textbf{\method}~(Ours) & \adepair{\underline{1.24}}{\underline{2.90}} & \textbf{7.94} \\
\bottomrule
\end{tabular}
\caption{WOD-E2E long-tail test results. ADE 3/5 denotes RFS-GT ADE at 3 and 5\,s. Bold and underlined values mark the best and second-best results.}
\label{tab:wod_e2e}
\end{table}

\subsection{Unified Prediction of Heterogeneous Futures}

The shared video expert, initialized from Wan2.2-5B and trained on all four streams, predicts RGB, segmentation, relative depth, and instance-track futures over 4\,s. Under matched settings, the multi-stream model achieves higher RGB PSNR and SSIM than the RGB-only model at all eight horizons. The multi-stream model achieves a mean PSNR of 19.55\,dB and a mean SSIM of 0.5844, whereas the RGB-only model achieves 19.44\,dB and 0.5820. The multi-stream model therefore shows no reduction in the reported RGB metrics, although the small differences do not establish an improvement. The supplementary material reports the horizon-wise curves for both settings.

The initialization ablation fixes the architecture, data, objectives, and optimization schedule to isolate the effect of video-expert initialization (Table~\ref{tab:ablation_pretraining}). Wan2.2-5B initialization yields better point estimates for all five metrics, including a 14.5-point increase in mIoU and a 7.3-point reduction in AbsRel relative to random initialization.

\begin{table}[!t]
\centering
\small
\setlength{\tabcolsep}{2pt}
\begin{tabular*}{\columnwidth}{@{\extracolsep{\fill}}lccccc@{}}
\toprule
Init. & PSNR$\uparrow$ & SSIM$\uparrow$ & mIoU$\uparrow$ & AbsRel$\downarrow$ & AssA@50$\uparrow$ \\
\midrule
Random & 18.51 & 0.5432 & 49.7 & 26.8 & 81.4 \\
\textbf{Wan2.2-5B} & \textbf{19.55} & \textbf{0.5844} & \textbf{64.2} & \textbf{19.5} & \textbf{84.4} \\
\bottomrule
\end{tabular*}
\caption{Effect of video-generative initialization under matched settings. Metrics are unweighted means over their evaluation horizons. AbsRel is reported in percent.}
\label{tab:ablation_pretraining}
\end{table}

Table~\ref{tab:structured_future_quality} compares native generation with Generate-then-Perceive. Native generation achieves a higher \(\delta_1\) (77.04 versus 70.99) and a lower AbsRel (19.49 versus 22.13). Generate-then-Perceive achieves higher mIoU (66.81 versus 64.22) and AssA@50 (86.88 versus 84.39). The horizon-wise results in the supplementary material preserve the same ordering. \method directly predicts segmentation, relative depth, and instance-track futures with one shared video expert and no stream-specific visual prediction heads.

\begin{figure*}[!t]
    \centering
    \includegraphics[width=\linewidth]{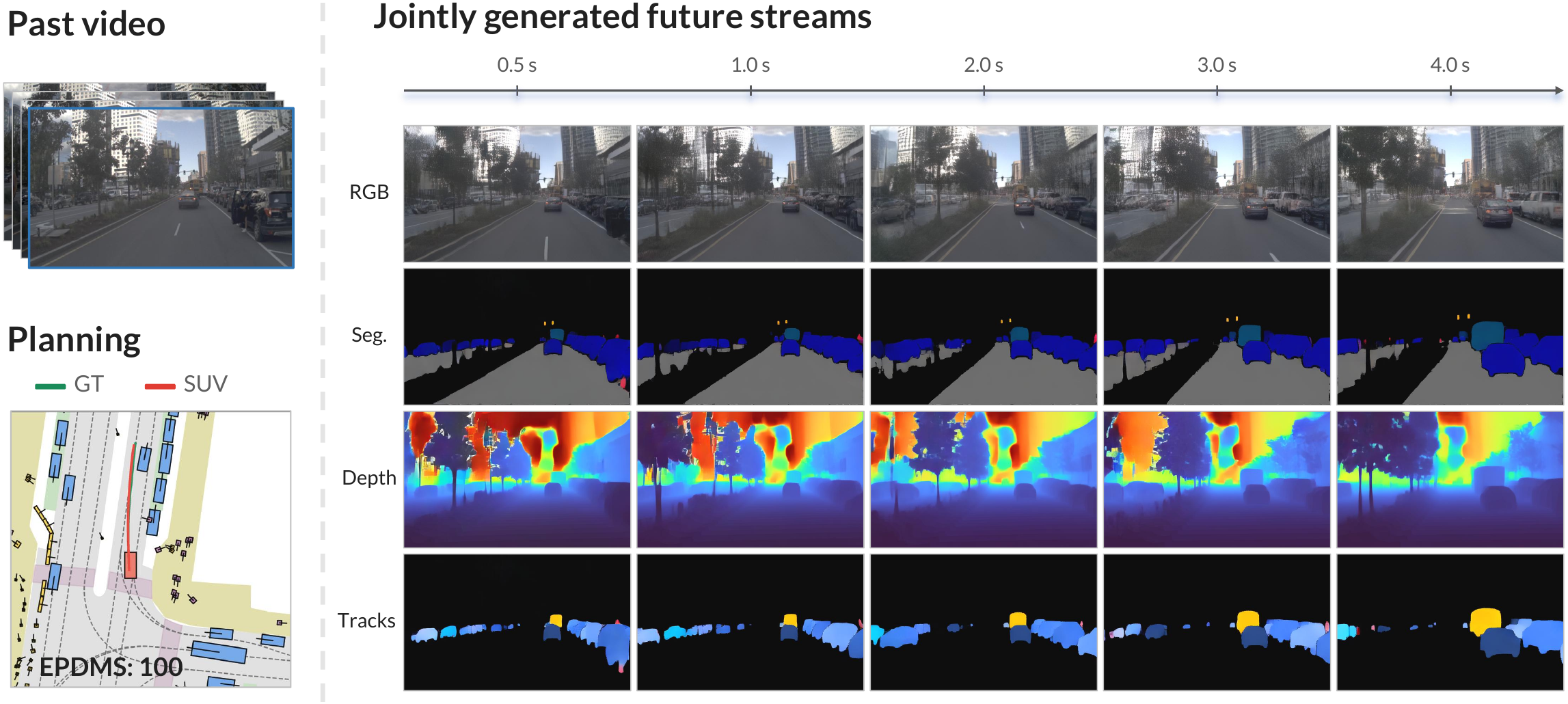}
    \caption{Qualitative multi-stream prediction and planning in a lane-change scenario. Left: Past video and planning. Right: The generated future streams have clear visual content, road layout, depth maps and consistent vehicle IDs across 4 seconds.}
    \label{fig:qualitative_results}
\end{figure*}

Figure~\ref{fig:qualitative_results} shows the four jointly generated future streams in a lane-change scenario. The road layout and vehicle locations remain consistent across the four streams. The planned trajectory closely matches the ground-truth trajectory and achieves an EPDMS of 100. The supplementary material provides additional horizon-wise curves and qualitative cases.

\begin{table}[!t]
\centering
\small
\setlength{\tabcolsep}{1.5pt}
\begin{tabular*}{\columnwidth}{@{\extracolsep{\fill}}lcccc@{}}
\toprule
Prediction route & mIoU$\uparrow$ & $\delta_1\uparrow$ & AbsRel$\downarrow$ & AssA@50$\uparrow$ \\
\midrule
Generate-then-Perceive & \textbf{66.81} & 70.99 & 22.13 & \textbf{86.88} \\
\textbf{Native Generation}~(\method) & 64.22 & \textbf{77.04} & \textbf{19.49} & 84.39 \\
\bottomrule
\end{tabular*}
\caption{Direct versus post-hoc structured future prediction. All metrics use teacher-generated references. The mIoU, \(\delta_1\), and AbsRel values are unweighted means over 0.5--4.0\,s. The AssA@50 values are unweighted means over 1.0--4.0\,s.}
\label{tab:structured_future_quality}
\end{table}

\subsection{Planning Benefits of Future Representations}

Table~\ref{tab:ablation_core} evaluates structured future supervision and future-stream access in a \(2\times2\) design. All four configurations use RGB future supervision, and the first row is the baseline without S/G/I supervision or future-stream access. S/G/I adds segmentation, relative-depth, and instance-track objectives. Access allows the action expert to read the RGB stream when S/G/I is absent and all four future streams when S/G/I is present. The four rows therefore represent the baseline, Access alone, S/G/I alone, and their combination. The S/G/I-only configuration is \methodwoaccess. It retains all four training objectives and action-prompt cross-attention but prevents the action expert from reading future-stream tokens. Comparing the two no-access rows isolates structured future supervision. S/G/I raises EPDMS from 89.7 to 90.7 on navtest and from 30.5 to 32.8 on navhard. Comparing the two S/G/I rows isolates future-stream access. Access raises EPDMS from 90.7 to 91.0 on navtest and from 32.8 to 36.9 on navhard. The corresponding Stage~1 and Stage~2 scores increase by 2.5 and 2.4 points. Under RGB-only supervision, Access raises EPDMS from 89.7 to 90.6 on navtest and from 30.5 to 35.0 on navhard.

Table~\ref{tab:ablation_modalities} isolates access to each structured future stream. Each ablated variant retains all four training objectives and RGB future-stream access while removing access to one structured stream. Removing segmentation, relative-depth, or instance-track access lowers navhard EPDMS from 36.9 to 36.4, 35.8, and 35.4, respectively. All three variants remain within 0.2 points of the full configuration on navtest. Relative-depth removal causes the largest Stage~1 decrease of 1.1 points, while instance-track removal causes the largest Stage~2 decrease of 1.3 points. Together, the ablations show that structured future supervision improves the RGB-supervised no-access baseline and that direct future-stream access provides further gains, particularly on navhard.

\begin{table}[!t]
\centering
\small
\setlength{\tabcolsep}{2.5pt}
\begin{tabular*}{\linewidth}{@{\extracolsep{\fill}}cc|c|ccc@{}}
\toprule
\multirow{2}{*}{S/G/I} & \multirow{2}{*}{Access} & navtest & \multicolumn{3}{c}{navhard} \\
\cmidrule(lr){3-3}\cmidrule(lr){4-6}
& & EPDMS$\uparrow$ & S1$\uparrow$ & S2$\uparrow$ & EPDMS$\uparrow$ \\
\midrule
\xmark & \xmark & 89.7 & 77.0 & 39.9 & 30.5 \\
\xmark & \cmark
& 90.6 & 80.0 & 43.8 & 35.0 \\
\cmark & \xmark
& 90.7 & 79.8 & 41.5 & 32.8 \\
\cmark & \cmark
& \textbf{91.0} & \textbf{82.3} & \textbf{43.9} & \textbf{36.9} \\
\bottomrule
\end{tabular*}
\caption{Structured future supervision and future-stream access. All rows use RGB future supervision, and S/G/I denotes segmentation, relative-depth, and instance-track supervision.}
\label{tab:ablation_core}

\setlength{\tabcolsep}{2pt}
\begin{tabular*}{\linewidth}{@{\extracolsep{\fill}}l|c|ccc@{}}
\toprule
\multirow{2}{*}{Variant} & navtest & \multicolumn{3}{c}{navhard} \\
\cmidrule(lr){2-2}\cmidrule(lr){3-5}
& EPDMS$\uparrow$ & S1$\uparrow$ & S2$\uparrow$ & EPDMS$\uparrow$ \\
\midrule
Full & \textbf{91.0} & 82.3 & \textbf{43.9} & \textbf{36.9} \\
\midrule
w/o Seg. Access
& 90.9 & \textbf{82.6} & 43.2 & 36.4 \\[1pt]
w/o Depth Access
& 90.8 & 81.2 & 43.9 & 35.8 \\[1pt]
w/o Track Access
& 90.9 & 82.0 & 42.6 & 35.4 \\
\bottomrule
\end{tabular*}
\caption{Access to individual structured future streams. All variants retain the four training objectives and RGB future-stream access. S1 and S2 denote the two navhard stages.}
\label{tab:ablation_modalities}
\end{table}

\subsection{Accuracy--Latency Trade-off}

\begin{table}[!ht]
\centering
\small
\setlength{\tabcolsep}{1.5pt}
\begin{tabular*}{\columnwidth}{@{\extracolsep{\fill}}lccc|cc@{}}
\toprule
Model & Steps & navtest$\uparrow$ & navhard$\uparrow$ & ms$\downarrow$ & Hz$\uparrow$ \\
\midrule
DriveVLA-W0 & -- & 86.9 & 24.4 & 690 & 1.45 \\
\midrule
\multirow{3}{*}{\method}
& 1  & 89.8 & 33.0 & \textbf{177} & \textbf{5.65} \\
& 2  & \textbf{91.0} & 36.1 & 288 & 3.48 \\
& 10 & \textbf{91.0} & \textbf{36.9} & 1356 & 0.74 \\
\bottomrule
\end{tabular*}
\caption{Accuracy--latency comparison. EPDMS follows the NAVSIM-v2 protocol. Each latency value is the mean of 500 runs on a single GeForce RTX 4090.}
\label{tab:ablation_steps}
\end{table}

Table~\ref{tab:ablation_steps} compares \method's solver settings with DriveVLA-W0 under the same GeForce RTX 4090 timing protocol. With one solver step, \method reaches 89.8 EPDMS on navtest and 33.0 on navhard at 177\,ms, exceeding DriveVLA-W0 by 2.9 and 8.6 points while reducing latency from 690 to 177\,ms, a 3.9$\times$ speedup. Two steps raise the scores to 91.0 and 36.1 at 288\,ms, remaining 2.4$\times$ faster than DriveVLA-W0. Ten steps yield the highest navhard score of 36.9 at 1356\,ms. The solver therefore exposes a controllable accuracy--latency range, with the one- and two-step settings improving both planning scores and latency over DriveVLA-W0 on this GPU.


\section{Conclusion}
We introduced \method, a unified driving framework that models RGB, segmentation, relative depth, and instance-track futures as video streams using a shared video expert without stream-specific visual prediction heads. A separate action expert attends to their latent tokens to generate the ego trajectory. Using a single front camera, \method outperforms a broad set of recent state-of-the-art methods on both NAVSIM-v2 splits, achieving 91.0 EPDMS on navtest and 36.9 on navhard. On the long-tail WOD-E2E benchmark, \method achieves a competitive RFS of 7.94. Controlled ablations show that Wan2.2-5B initialization yields higher future-prediction point estimates than random initialization. Structured supervision raises planning point estimates without future-stream access, while direct access provides additional gains, particularly on navhard. These results support video generation as a shared future-scene predictor for planning.

\FloatBarrier
\begingroup
\small
\bibliography{aaai2027,supplement_refs}
\endgroup

\normalfont\normalsize
\setcounter{secnumdepth}{2}
\setcounter{section}{0}
\setcounter{subsection}{0}
\setcounter{figure}{0}
\setcounter{table}{0}
\setcounter{equation}{0}

\clearpage
\twocolumn[
\vbox to 2.25in{
\hsize\textwidth
\linewidth\hsize
\vskip 0.625in minus 0.125in
\centering
{\LARGE\bfseries SUV: Future Scene Understanding as Video Generation for End-to-End Driving\par}
\vskip 0.1in plus 0.5fil minus 0.05in
{\Large\bfseries Supplementary Material\par}
\vskip 0.2em plus 0.25fil
\vskip 1em plus 2fil
}
]

\noindent This supplement defines the planning and future-scene metrics used in the main paper (Section~\ref{sec:supp_evaluation_protocols}). Section~\ref{sec:supp_additional_evidence} presents additional planning results and qualitative examples. Section~\ref{sec:supp_future_scene_prediction} reports horizon-wise future-scene results. Section~\ref{sec:supp_reproducibility} documents structured targets, benchmark configuration, training, and inference.

\section{Evaluation Metrics}
\label{sec:supp_evaluation_protocols}

\subsection{Planning Metrics}
\label{sec:supp_planning_metrics}

All NAVSIM component scores lie in \([0,1]\), and leaderboard scores are split means multiplied by 100. WOD-E2E reports RFS on a \([0,10]\) scale.

\paragraph{NAVSIM PDMS and EPDMS.}
PDMS and EPDMS combine multiplicative component scores with a weighted average of the remaining terms~\cite{navsim,navsimv2}. For \(x_c\in[0,1]\), let \(\mathcal P\) and \(\mathcal B\) denote these two groups:
\begin{equation}
\mathrm{Score}=\left(\prod_{c\in\mathcal P}x_c\right)
\times
\frac{\sum_{c\in\mathcal B}w_c\times x_c}
{\sum_{c\in\mathcal B}w_c}.
\label{eq:supp_navsim_score}
\end{equation}
A zero multiplier makes \(\mathrm{Score}=0\).

\begin{center}
\begin{minipage}{\columnwidth}
\centering
\footnotesize
\setlength{\tabcolsep}{3pt}
\renewcommand{\arraystretch}{1.02}
\begin{tabular}{@{}lp{0.47\columnwidth}cc@{}}
\toprule
Term & Meaning & PDMS & EPDMS \\
\midrule
NC    & No at-fault collisions       & M & M \\
DAC   & Drivable-area compliance     & M & M \\
DDC   & Driving-direction compliance & -- & M \\
TLC   & Traffic-light compliance     & -- & M \\
TTC   & Time to collision            & 5 & 5 \\
EP    & Ego progress                 & 5 & 5 \\
C     & Comfort                      & 2 & -- \\
LK    & Lane keeping                 & -- & 2 \\
HC    & History comfort              & -- & 2 \\
EC    & Extended comfort             & -- & 2 \\
\bottomrule
\end{tabular}
\captionof{table}{PDMS and EPDMS component roles and weights. M marks multipliers, and dashes mark absent components.}
\label{tab:supp_pdms_epdms_weights}
\end{minipage}
\end{center}

For EPDMS, \(x_c\) is the effective component score after human-penalty filtering. The evaluator applies this filter to NC, DAC, DDC, TLC, EP, TTC, LK, and HC in original scenes. If the corresponding human score is 0, the component is set to 1. Synthetic follow-ups and EC are evaluated without this filter. The weighted-term denominator is 12 for PDMS, 16 for EPDMS with EC, and 14 when EC is unavailable.

\paragraph{NAVSIM-v2 navhard.}
Navhard applies the EPDMS defined above within a two-stage pseudo-simulation protocol~\cite{navsimv2}. Stage-1 evaluates an initial 4-s plan on a logged observation, whereas Stage-2 evaluates new 4-s plans on associated 3DGS follow-up observations. Each follow-up represents a plausible outcome of a different precomputed Stage-1 rollout. This protocol retains offline evaluation while testing planning after an observation shift rather than a continuous 8-s rollout.

Let \(\mathcal G\) denote the set of Stage-1 scenes. For \(g\in\mathcal G\), \(s_g^{(1)}\in[0,1]\) is its unscaled EPDMS and \(\mathbf e_g=(e_g^x,e_g^y)\) is its predicted rear-axle endpoint. For follow-up \(k\), \(s_{gk}^{(2)}\in[0,1]\) is its unscaled EPDMS and \(\mathbf b_{gk}=(b_{gk}^x,b_{gk}^y)\) is its rear-axle start position. The evaluator assigns larger normalized Gaussian weights to follow-ups whose start positions are closer to \(\mathbf e_g\):
\(\alpha_{gk}\propto\exp[-\|\mathbf e_g-\mathbf b_{gk}\|_2^2/(2\times\sigma^2)]\), where \(\sum_k\alpha_{gk}=1\) and \(\sigma^2=0.1\,\mathrm{m}^2\). The combined score is
\begin{equation}
\begin{aligned}
\mathrm{EPDMS}_{\mathrm{hard}}
&=\frac{100}{|\mathcal G|}\times\sum_{g\in\mathcal G}
s_g^{(1)}\times\left(\sum_k\alpha_{gk}\times s_{gk}^{(2)}\right).
\end{aligned}
\label{eq:supp_navhard}
\end{equation}
The inner sum estimates Stage-2 performance near the predicted Stage-1 endpoint. Its product with \(s_g^{(1)}\) requires both plans to score well. The outer mean aggregates all Stage-1 scenes, and the factor \(100\) converts the unscaled mean to the leaderboard scale.

\paragraph{WOD-E2E RFS.}
Rater Feedback Score (RFS) compares a predicted 5-s trajectory with three human-rated references, each assigned a score \(s_p\in[0,10]\)~\cite{wode2e}. At each evaluation time \(t\in\{3,5\}\)\,s, the absolute longitudinal and lateral errors, measured in the local frame of reference trajectory \(p\), are normalized by initial-speed-dependent trust-region thresholds:
\begin{equation}
\begin{aligned}
d_{p,t}
&=\max\left(
\frac{\Delta_{\mathrm{lng},p,t}}{\tau_{\mathrm{lng},t}},
\frac{\Delta_{\mathrm{lat},p,t}}{\tau_{\mathrm{lat},t}}
\right),\\
r_{p,t}
&=s_p\times0.1^{\max(d_{p,t}-1,0)},\\
\widetilde{\mathrm{RFS}}
&=\frac{1}{2}\times\sum_{t\in\{3,5\}}
\max_{p\in\{1,2,3\}}r_{p,t}.
\end{aligned}
\label{eq:supp_rfs}
\end{equation}
A prediction inside a trust region retains \(s_p\), and its score decays exponentially outside the region. The maximum is taken independently over references at 3 and 5\,s, and \(\widetilde{\mathrm{RFS}}\) averages the two values. The floor applies when no single reference satisfies both \(d_{p,3}\leq1\) and \(d_{p,5}\leq1\). In this case, \(\mathrm{RFS}=\max(4,\widetilde{\mathrm{RFS}})\). Otherwise, \(\mathrm{RFS}=\widetilde{\mathrm{RFS}}\). Dataset-level RFS is the mean score across the 11 scenario clusters. RFS-GT ADE reports 3- and 5-s displacement from the highest-rated recorded reference, averaged over test frames.

\subsection{Future-Scene Metrics}
\label{sec:supp_structured_evaluation}

The future-scene metrics below are evaluated on NAVSIM-v2 navtest. RGB metrics compare predicted and recorded future frames. Structured metrics compare decoded predictions with frozen-teacher targets derived from recorded future RGB. Section~\ref{sec:supp_structured_targets} details target encoding and decoding.

\begin{center}
\begin{minipage}{\columnwidth}
\centering
\small
\setlength{\tabcolsep}{4pt}
\renewcommand{\arraystretch}{1.08}
\begin{tabular}{@{}l>{\raggedright\arraybackslash}p{0.70\columnwidth}@{}}
\toprule
Metric & Meaning \\
\midrule
PSNR$\uparrow$ & Pixel fidelity on a logarithmic scale. \\
SSIM$\uparrow$ & Structural similarity between RGB frames. \\
mIoU$\uparrow$ & Mean semantic overlap across evaluated classes. \\
\(\delta_1\uparrow\) & Fraction of depth pixels within a factor of 1.25. \\
AbsRel$\downarrow$ & Mean absolute relative depth error. \\
AssA@50$\uparrow$ & Track-association accuracy at a mask-IoU threshold of 0.50. \\
\bottomrule
\end{tabular}
\captionof{table}{Future-scene evaluation metrics. Arrows indicate the preferred direction.}
\label{tab:supp_future_scene_metrics}
\end{minipage}
\end{center}

PSNR and SSIM use all image pixels and are averaged equally over clips at each horizon. Relative depth metrics cover the full image without the 50-m restriction and use the valid percentile mask \(\mathcal V_{c,h}\) defined below. To focus on driving-relevant content, mIoU and AssA@50 evaluate Car, Truck, Bus, Bicycle, and Pedestrian within a \(50\,\mathrm{m}\) ego-centric region. At each horizon, segmentation predictions are decoded to the nearest palette color, and mIoU is computed from a confusion matrix pooled over the split. Classes with zero union are excluded.

For relative depth, \(c\) indexes a clip and \(h\in\{1,\ldots,8\}\) indexes a future frame. Let \(q_{0.01,c}\) and \(q_{0.99,c}\) denote the first and 99th percentiles of the finite, positive DA3-reference depths in clip \(c\). The set \(\mathcal V_{c,h}\) contains full-image pixels whose reference depths lie within this interval. We estimate one scale and shift jointly from all eight frames:
\begin{equation}
\begin{aligned}
(a_c,b_c)
&=\arg\min_{a,b}\sum_{h=1}^{8}\sum_{p\in\mathcal V_{c,h}}
\left(a\widehat d_{c,h}(p)+b-d_{c,h}(p)\right)^2,\\
\widetilde d_{c,h}(p)
&=\operatorname{clip}_{[q_{0.01,c},q_{0.99,c}]}\!\left(
a_c\widehat d_{c,h}(p)+b_c
\right),
\end{aligned}
\label{eq:supp_depth_alignment}
\end{equation}
where \(\widehat d\) denotes the decoded normalized depth prediction and \(d\) denotes the DA3-reference relative depth. The same \((a_c,b_c)\) and clipping interval are used at every horizon in clip \(c\). We then compute
\begin{equation}
\begin{aligned}
\mathrm{AbsRel}_{c,h}
&=\frac{1}{|\mathcal V_{c,h}|}
\sum_{p\in\mathcal V_{c,h}}
\frac{|\widetilde d_{c,h}(p)-d_{c,h}(p)|}{d_{c,h}(p)},\\
\rho_{c,h}(p)
&=\max\left(
\frac{\widetilde d_{c,h}(p)}{d_{c,h}(p)},
\frac{d_{c,h}(p)}{\widetilde d_{c,h}(p)}
\right),\\
\delta_{1,c,h}
&=\frac{1}{|\mathcal V_{c,h}|}
\sum_{p\in\mathcal V_{c,h}}
\mathbf{1}\!\left[\rho_{c,h}(p)<1.25\right].
\end{aligned}
\label{eq:supp_depth_metrics}
\end{equation}
We report \(\delta_1\) and AbsRel as affine-aligned, percentile-clipped measures of agreement with the DA3 teacher. At each horizon, both metrics are averaged equally over clips.

AssA@50 is the association-accuracy component of HOTA~\cite{luiten2021hota_supp}, evaluated at a mask-IoU threshold of 0.50. Within clip \(c\) and cumulative prefix \(1{:}h\), each matched detection \(m\in\mathrm{TP}_{c,h}\) receives the per-match association score
\begin{equation}
\begin{aligned}
a(m)
&=\frac{|\mathrm{TPA}(m)|}
{|\mathrm{TPA}(m)|+|\mathrm{FPA}(m)|+|\mathrm{FNA}(m)|},\\
A_h
&=\frac{\sum_c\sum_{m\in\mathrm{TP}_{c,h}}a(m)}
{\sum_c|\mathrm{TP}_{c,h}|},
\qquad
\overline A=\frac{1}{7}\sum_{h=2}^{8}A_h.
\end{aligned}
\label{eq:supp_assa}
\end{equation}
Here, \(c\) indexes clips, and \(\mathrm{TP}_{c,h}\) contains accepted matches pooled over the five evaluated classes. TPA, FPA, and FNA are association true positives, false positives, and false negatives for the predicted--reference identity pair containing \(m\). This definition keeps identities clip-local and micro-averages matched detections across the split. We evaluate prefixes \(h=2,\ldots,8\). The single-frame prefix provides no temporal association evidence and is therefore excluded. The reported RGB, segmentation, and depth scores are unweighted means over eight horizons, whereas AssA@50 is the unweighted mean over seven horizons. We report mIoU, \(\delta_1\), AbsRel, and AssA@50 as percentages.

\FloatBarrier

\section{Additional Planning Results}
\label{sec:supp_additional_evidence}

\subsection{NAVSIM-v1 Planning Results}
\label{sec:supp_navsim_v1}

Table~\ref{tab:supp_navsim_v1} reports NAVSIM-v1 navtest performance under the official PDMS protocol~\cite{navsim}. We take baseline values from published results for traditional E2E policies~\cite{transfuser,diffusiondrive,world4drive,hydramdppp,seerdrive,drivejepa} and VLA-based policies~\cite{autovla,drivevlaw0,sgdrive,autodrivep3,recogdrive}. The WAM-based baselines follow published reports~\cite{imagidrive,pwm,drivelaw,eponav2,metis}. With a single front camera, \method achieves 90.8 PDMS, matching ReCogDrive for the highest score among the listed learned methods.

\begin{center}
\centering
\small
\setlength{\tabcolsep}{1.2pt}
\renewcommand{\arraystretch}{1.05}
\begin{tabular*}{\columnwidth}{@{\extracolsep{\fill}}lcccccc@{}}
\toprule
Method & NC$\uparrow$ & DAC$\uparrow$ & TTC$\uparrow$ & Comf.$\uparrow$ & EP$\uparrow$ & PDMS$\uparrow$ \\
\midrule
Human & 100 & 100 & 100 & 99.9 & 87.5 & 94.8 \\
\midrule
\rowcolor{groupgray}
\multicolumn{7}{c}{\textit{Traditional E2E policies}} \\
TransFuser & 97.7 & 92.8 & 92.8 & \textbf{100} & 79.2 & 84.0 \\
DiffusionDrive & 98.2 & 96.2 & 94.7 & \textbf{100} & 82.2 & 88.1 \\
World4Drive & 97.4 & 94.3 & 92.8 & \textbf{100} & 79.9 & 85.1 \\
Hydra-MDP++ & 97.6 & 96.0 & 93.1 & \textbf{100} & 80.4 & 86.6 \\
SeerDrive & 98.4 & 97.0 & 94.9 & \underline{99.9} & 83.2 & 88.9 \\
Drive-JEPA & 98.7 & 96.2 & \textbf{100} & 95.5 & 82.9 & 89.0 \\
\midrule
\rowcolor{groupgray}
\multicolumn{7}{c}{\textit{VLA-based policies}} \\
AutoVLA & 98.4 & 95.6 & \underline{98.0} & \underline{99.9} & 81.9 & 89.1 \\
DriveVLA-W0 & 98.7 & 96.2 & 95.5 & \textbf{100} & 82.2 & 88.4 \\
SGDrive-IL & 98.6 & 95.1 & 95.4 & \textbf{100} & 81.2 & 87.4 \\
AutoDrive-P$^3$ & \textbf{99.1} & 97.4 & 96.5 & \textbf{100} & \underline{84.8} & \underline{90.6} \\
ReCogDrive & 97.9 & 97.3 & 94.9 & \textbf{100} & \textbf{87.3} & \textbf{90.8} \\
\midrule
\rowcolor{groupgray}
\multicolumn{7}{c}{\textit{WAM-based policies}} \\
ImagiDrive & 98.6 & 96.2 & 94.5 & \textbf{100} & 80.5 & 87.4 \\
PWM & 98.6 & 95.9 & 95.4 & \textbf{100} & 81.8 & 88.1 \\
DriveLaW & \underline{99.0} & 97.1 & 96.7 & \textbf{100} & 81.3 & 89.1 \\
EponaV2 & 98.6 & \textbf{97.9} & 95.7 & \textbf{100} & \underline{84.8} & 90.4 \\
Metis & 98.3 & 97.1 & 94.7 & \textbf{100} & 83.4 & 89.1 \\
Metis~(Top 6) & 98.5 & 97.5 & 95.1 & \textbf{100} & 84.0 & 89.7 \\
\midrule
\textbf{\method} & \textbf{99.1} & \underline{97.8} & 96.7 & \textbf{100} & 84.6 & \textbf{90.8} \\
\bottomrule
\end{tabular*}
\captionof{table}{NAVSIM-v1 navtest planning under each method's published sensor and candidate configuration. Boldface and underlining indicate the best and second-best learned results. Metis~(Top 6) uses best-of-6 selection, while \method outputs one trajectory.}
\label{tab:supp_navsim_v1}
\end{center}

\subsection{Planning-Score Reproducibility and Scene-Level Variation}
\label{sec:supp_scene_variation}

We repeated the NAVSIM-v1 and NAVSIM-v2 navtest evaluations six times with fixed checkpoints and inference seed 42. Every repetition returned the same leaderboard-scale scores to two decimal places: 90.84 PDMS on NAVSIM-v1 navtest and 91.01 EPDMS on NAVSIM-v2 navtest. These repetitions assess fixed-checkpoint numerical reproducibility.

Table~\ref{tab:supp_scene_variation} reports the empirical mean and sample standard deviation across scene-level scores using an \(N-1\) denominator. Each navtest scene is one observation. Scene scores are multiplied by 100 before summarization, and the results are rounded to two decimal places. The standard deviation describes scene-level dispersion.

\begin{center}
\begin{minipage}{\columnwidth}
\centering
\footnotesize
\setlength{\tabcolsep}{3pt}
\renewcommand{\arraystretch}{1.08}
\begin{tabular*}{\columnwidth}{@{\extracolsep{\fill}}lcccc@{}}
\toprule
Benchmark & Metric & \(N\) & Mean & Scene SD \\
\midrule
NAVSIM-v1 navtest & PDMS  & 12{,}146 & 90.84 & 17.84 \\
NAVSIM-v2 navtest & EPDMS & 12{,}146 & 91.01 & 18.21 \\
\bottomrule
\end{tabular*}
\captionof{table}{Scene-level variation in the primary planning scores. Metric definitions are given in Section~\ref{sec:supp_planning_metrics}.}
\label{tab:supp_scene_variation}
\end{minipage}
\end{center}

\subsection{Qualitative Results on navtest}
\label{sec:supp_navsim_qualitative}

\begin{figure*}[p]
    \centering
    \includegraphics[width=0.92\textwidth,pagebox=cropbox]{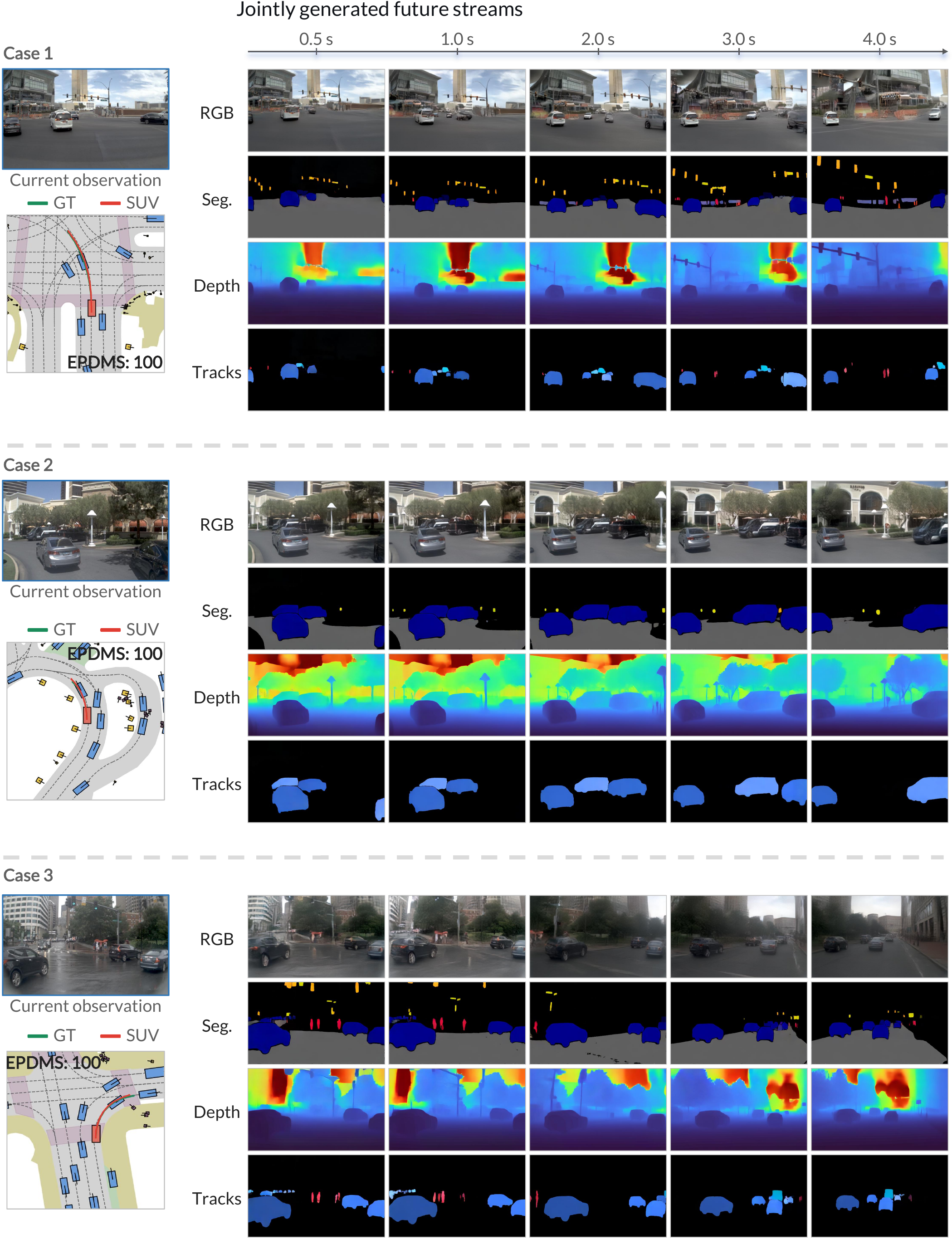}
    \caption{High-curvature turns on NAVSIM-v2 navtest. Each case shows the current observation, bird's-eye-view trajectories, and jointly generated RGB, semantic segmentation, relative depth, and instance tracks at five prediction horizons. Green and red denote the recorded human and \method trajectories.}
    \label{fig:supp_navsim_cases}
\end{figure*}

\begin{figure*}[t]
    \centering
    \includegraphics[width=0.92\textwidth,pagebox=cropbox]{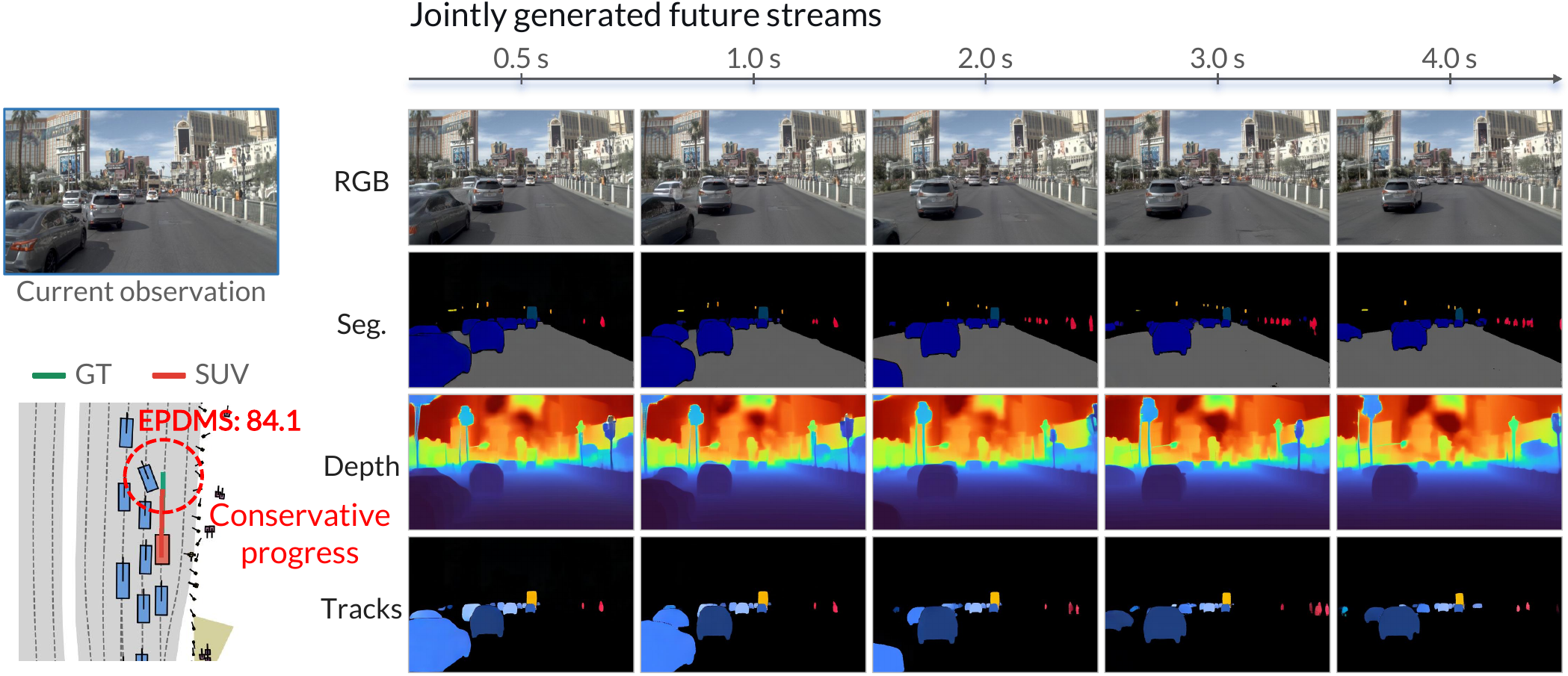}
    \caption{Lower progress in dense urban traffic. The \method trajectory follows the recorded human route but advances less over the 4-s horizon. The panels show jointly generated RGB, semantic segmentation, relative depth, and instance tracks. The case receives an EPDMS of 84.1, and the colors follow Figure~\ref{fig:supp_navsim_cases}.}
    \label{fig:supp_navsim_conservative}
\end{figure*}

Figures~\ref{fig:supp_navsim_cases} and~\ref{fig:supp_navsim_conservative} show complementary behaviors on NAVSIM-v2 navtest. Figure~\ref{fig:supp_navsim_cases} presents three high-curvature turns: a wide signalized intersection, a curved urban road with nearby traffic, and a wet-road turn. Each case pairs the current observation and bird's-eye-view trajectories with jointly generated RGB, semantic segmentation, relative depth, and instance tracks at 0.5, 1.0, 2.0, 3.0, and 4.0\,s. Across these turns, the generated streams retain the road geometry and nearby actors as the heading changes. Figure~\ref{fig:supp_navsim_conservative} presents a dense-traffic case in which the \method trajectory follows the recorded human route but advances less amid surrounding traffic. The resulting EPDMS is 84.1, reflecting lower progress.

\subsection{Qualitative Zero-Shot Transfer to In-House Data}
\label{sec:supp_cross_domain_example}

We apply \method without adaptation to two non-public in-house driving clips from a source domain outside the public benchmarks. Each clip contains a front-camera observation and eight future ego poses over 4\,s. The clips were excluded from training and model selection and serve as qualitative examples of cross-domain transfer. Figure~\ref{fig:supp_zeroshot_inhouse_planning} shows both results. The nighttime and daytime examples differ in road geometry and surrounding traffic, while both predicted trajectories follow the recorded lateral evolution and forward progress. This alignment persists throughout the 4-s horizon rather than appearing only at the final waypoint.
\par\medskip
\noindent\begin{minipage}{\linewidth}
    \centering
    \includegraphics[width=\linewidth,pagebox=cropbox]{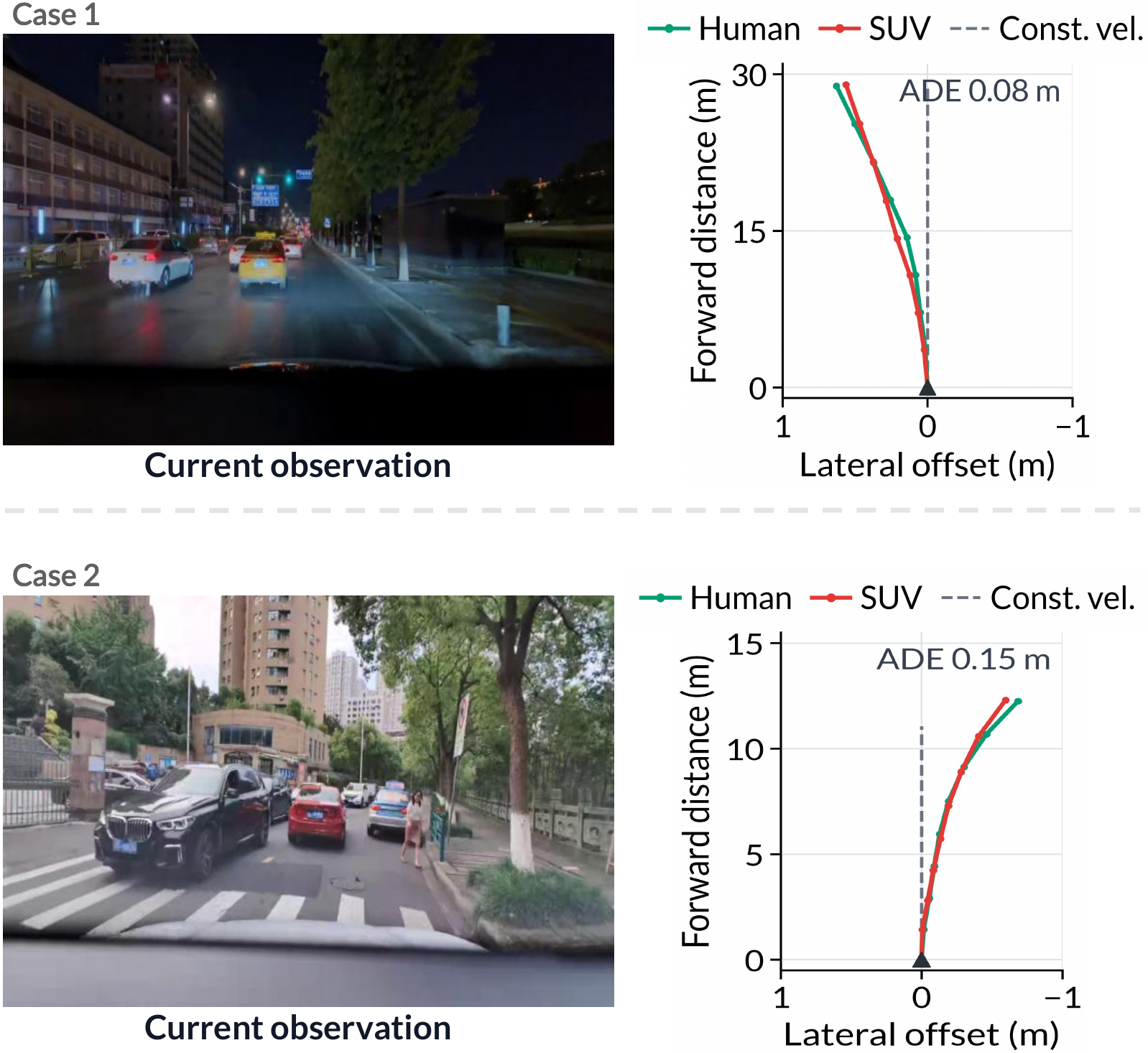}
    \captionof{figure}{Qualitative zero-shot planning on two in-house clips. Each row shows one example. Green, red, and dashed gray denote the recorded human, \method, and constant-velocity trajectories. The ADEs for the top and bottom examples are 0.08 and 0.15\,m over eight waypoints spanning 4\,s.}
    \label{fig:supp_zeroshot_inhouse_planning}
\end{minipage}

\FloatBarrier

\section{Future-Scene Prediction}
\label{sec:supp_future_scene_prediction}

\begin{figure}[!b]
    \centering
    \includegraphics[width=\columnwidth]{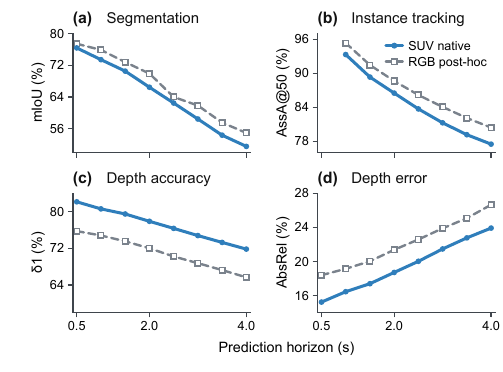}
    \caption{Horizon-wise teacher-agreement metrics for native generation (SUV native) and Generate-then-Perceive (RGB post-hoc). Semantic segmentation and relative depth cover 0.5 to 4.0\,s, and instance tracks cover 1.0 to 4.0\,s. Higher is better except for AbsRel.}
    \label{fig:supp_structured_future_quality}
\end{figure}

\begin{figure}[!t]
    \centering
    \includegraphics[width=\columnwidth]{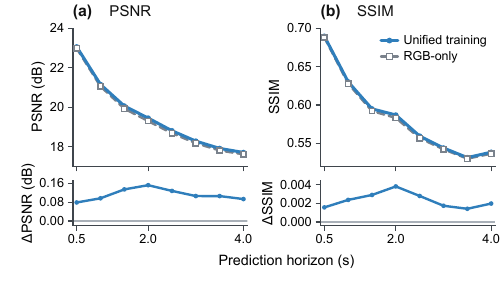}
    \caption{Horizon-wise RGB metrics for multi-stream (Unified) and RGB-only training. The lower panels plot Unified minus RGB-only. Values above zero indicate higher scores for Unified.}
    \label{fig:supp_rgb_future_quality}
\end{figure}

Figures~\ref{fig:supp_structured_future_quality} and~\ref{fig:supp_rgb_future_quality} report horizon-wise values under Section~\ref{sec:supp_structured_evaluation}. Figure~\ref{fig:supp_structured_future_quality} compares native generation (SUV native), which predicts the structured streams directly, with Generate-then-Perceive (RGB post-hoc), which applies SAM~3 and DA3 to generated RGB. Figure~\ref{fig:supp_rgb_future_quality} compares multi-stream training (Unified) with RGB-only training under matched settings.

All four structured metrics become less favorable with prediction horizon. The ordering in the main-paper averages holds at every horizon: native generation has higher \(\delta_1\) and lower AbsRel, while Generate-then-Perceive has higher mIoU and AssA@50. Direct generation is therefore competitive without stream-specific visual prediction heads.

RGB PSNR and SSIM decrease with prediction horizon for both models. Multi-stream training is slightly higher at all eight horizons, with the largest gaps near 2\,s. These small positive differences show no reduction in the reported RGB metrics but do not establish an improvement.

\section{Implementation and Reproducibility}
\label{sec:supp_reproducibility}

\subsection{Structured Target Construction and Decoding}
\label{sec:supp_structured_targets}

Frozen teachers convert each recorded future RGB clip into videos for semantic segmentation, relative depth, and instance tracks. The videos use the same \(640\times384\) camera grid and 2-Hz sampling rate. We denote the number of future video frames by \(T\), with \(T=8\) over 4\,s for NAVSIM and \(T=10\) over 5\,s for WOD-E2E. These videos serve as training targets for both benchmarks and as evaluation references for the NAVSIM future-scene analysis.

\paragraph{Semantic segmentation.}
\label{sec:supp_semantic_targets}

We apply SAM~3~\cite{carion2025sam3segmentconcepts} independently to each future frame. Its image processor directly resizes each frame to \(1008\times1008\) for inference. We use a confidence threshold of 0.5 and the 11 prompts listed in Table~\ref{tab:supp_semantic_palette}. For each prompt, we retain up to 32 detections in descending confidence order. Before applying the 128-detection frame cap, we place road detections first and preserve confidence order within the road and non-road groups. We resize masks to the target grid with nearest-neighbor interpolation.

The class map starts as background. Road detections form a fill layer and update only background pixels. A non-road detection can replace background or road but not another non-road label. We process non-road detections in descending confidence order, so the highest-confidence non-road mask takes priority where non-road masks overlap. The resulting class indices are rendered as uint8 RGB with the fixed palette in Table~\ref{tab:supp_semantic_palette}. Generated segmentation videos are decoded by mapping each pixel to the nearest fixed palette color.

\begin{center}
\begin{minipage}{\columnwidth}
\centering
\small
\setlength{\tabcolsep}{5pt}
\renewcommand{\arraystretch}{1.04}
\begin{tabular*}{\columnwidth}{@{\extracolsep{\fill}}lc@{}}
\toprule
Class & RGB value \\
\midrule
Background & \((0,0,0)\) \\
Road & \((96,96,96)\) \\
Car & \((0,0,142)\) \\
Truck & \((0,0,70)\) \\
Bus & \((0,60,100)\) \\
Pedestrian & \((220,20,60)\) \\
Bicycle & \((119,11,32)\) \\
Motorcycle & \((0,0,230)\) \\
Traffic light & \((250,170,30)\) \\
Traffic sign & \((220,220,0)\) \\
Traffic cone & \((255,80,0)\) \\
Barrier & \((102,102,156)\) \\
\bottomrule
\end{tabular*}
\captionof{table}{Semantic classes and fixed RGB values. Background denotes pixels not assigned to a prompt.}
\label{tab:supp_semantic_palette}
\end{minipage}
\end{center}

\paragraph{Relative depth.}
\label{sec:supp_depth_targets}

We run DA3-LARGE~\cite{depthanything3} jointly on all \(T\) future frames using its default upper-bound resize, which limits the longest input side to 504 pixels. We then resize the relative depth maps bilinearly to the target grid. Let \(D\) contain the finite, positive DA3 outputs in one clip, and let \(Q_p(D)\) denote its \(p\)-th quantile. We set
\begin{equation}
\ell=Q_{0.01}(D),
\qquad
u=Q_{0.99}(D),
\label{eq:supp_depth_bounds}
\end{equation}
and replace \(u\) by \(\ell+1\) if \(u-\ell<10^{-6}\). Each finite, positive value \(d\) is encoded as
\begin{equation}
\bar d=\operatorname{clip}\!\left(\frac{d-\ell}{u-\ell},0,1\right),
\qquad
k=\operatorname{round}(255\bar d).
\label{eq:supp_depth_quantization}
\end{equation}
Nonfinite or nonpositive values receive \(\bar d=0\). The integer \(k\in\{0,\ldots,255\}\) selects one entry from the 256-color Google Turbo lookup table. Shared clip-level bounds give every frame the same monotonic depth-to-LUT-index mapping. Values outside \([\ell,u]\) saturate at the endpoint colors.

At evaluation, each generated depth pixel is mapped to its nearest Turbo entry. If \(\widehat k\) is the selected LUT index, the decoder returns \(\widehat{\bar d}=\widehat k/255\). For NAVSIM future-scene evaluation, Equation~\ref{eq:supp_depth_alignment} fits one affine map from these normalized values to the DA3-reference scale over all eight frames.

\paragraph{Instance tracks.}
\label{sec:supp_instance_targets}

The instance stream encodes semantic class and clip-local identity in a single RGB video. We process the six class prompts sequentially in one \(T\)-frame SAM~3 video session. Each prompt is introduced at the first future frame and propagated through the clip. Table~\ref{tab:supp_instance_palette} lists the prompts and their base colors. We threshold output probabilities at 0.5. Within each frame, masks are composited from largest to smallest, allowing smaller traffic participants to overwrite larger masks.

We encode class-local SAM~3 identity \(j\in\{0,1,\ldots\}\) using a zero-indexed vector with components \(\Delta_0,\ldots,\Delta_6\):
\begin{equation}
\begin{aligned}
\boldsymbol{\Delta}
&=(0,0.18,-0.18,0.34,-0.34,0.50,-0.50),\\
\delta_j&=\Delta_{j\bmod 7}.
\end{aligned}
\label{eq:supp_instance_shades}
\end{equation}
Its component-wise RGB value is
\begin{equation}
\mathbf{p}_{c,j}=
\begin{cases}
\operatorname{round}\!\left(\mathbf{b}_c+(255\mathbf{1}-\mathbf{b}_c)\delta_j\right), & \delta_j\geq0,\\
\operatorname{round}\!\left(\mathbf{b}_c(1+\delta_j)\right), & \delta_j<0,
\end{cases}
\label{eq:supp_instance_color}
\end{equation}
where \(\mathbf{1}=(1,1,1)\), and each channel is clipped to \([0,255]\). The color stays fixed for the entire track. Background pixels are black.

During decoding, pixels with Euclidean RGB distance greater than 30 from black are treated as foreground and mapped to the nearest of the 42 class-brightness codes. We extract 8-connected components independently for each code and remove components smaller than 32 pixels. Within each code, one-to-one Hungarian assignment links components in adjacent frames. The cost equals one minus mask IoU plus centroid displacement normalized by the image diagonal, and the returned assignments define the track links. Unmatched current components initialize new clip-local tracks, while unmatched previous tracks terminate. Tracks are linked only across consecutive frames. For NAVSIM native-generation evaluation, this decoder is applied to both generated and reference instance videos. The instance stream encodes all six prompted classes, whereas AssA@50 uses the five-class evaluation scope defined in Section~\ref{sec:supp_structured_evaluation}.

\begin{center}
\begin{minipage}{\columnwidth}
\centering
\small
\setlength{\tabcolsep}{5pt}
\renewcommand{\arraystretch}{1.05}
\begin{tabular*}{\columnwidth}{@{\extracolsep{\fill}}lc@{}}
\toprule
Prompt & Base RGB value \\
\midrule
Car & \((64,128,255)\) \\
Truck & \((0,200,255)\) \\
Bus & \((255,190,0)\) \\
Pedestrian & \((255,64,96)\) \\
Bicycle & \((80,220,100)\) \\
Motorcycle & \((220,90,255)\) \\
\bottomrule
\end{tabular*}
\captionof{table}{Instance-track base colors. Seven brightness offsets encode the class-local track ID modulo seven.}
\label{tab:supp_instance_palette}
\end{minipage}
\end{center}

\FloatBarrier

\subsection{Benchmark-Specific Configurations}
\label{sec:supp_benchmark_configuration}

WOD-E2E uses recorded future front-camera frames as RGB targets. Its targets for semantic segmentation, relative depth, and instance tracks follow Section~\ref{sec:supp_structured_targets} and share the same ten timestamps. A separately trained WOD-E2E model jointly denoises the four future streams and action trajectory. Its action tokens attend to all latent tokens from every future stream.

\begin{center}
\begin{minipage}{\columnwidth}
\centering
\footnotesize
\setlength{\tabcolsep}{3pt}
\renewcommand{\arraystretch}{1.08}
\begin{tabular}{@{}>{\raggedright\arraybackslash}p{0.21\columnwidth}>{\raggedright\arraybackslash}p{0.34\columnwidth}>{\raggedright\arraybackslash}p{0.34\columnwidth}@{}}
\toprule
Setting & NAVSIM & WOD-E2E \\
\midrule
Observation frames & 4 frames, 2\,Hz & 5 frames, 2\,Hz \\
Future streams & 8 frames each, 2\,Hz, 4\,s & 10 frames each, 2\,Hz, 5\,s \\
Trajectory & 8 points, 2\,Hz, 4\,s & 20 points, 4\,Hz, 5\,s \\
Generated streams & RGB, segmentation, relative depth, instance tracks & RGB, segmentation, relative depth, instance tracks \\
Image size & \(640\times384\) & \(640\times384\) \\
Candidates & One trajectory & One trajectory \\
\bottomrule
\end{tabular}
\captionof{table}{Temporal input and output configurations for NAVSIM and WOD-E2E.}
\label{tab:supp_benchmark_configuration}
\end{minipage}
\end{center}

The WOD-E2E preprocessing crops each front-camera frame to the \(5{:}3\) target aspect ratio and then bilinearly resizes it to \(640\times384\). WOD-E2E evaluation uses the global-step-100{,}000 checkpoint with bfloat16 precision and 10 Euler steps on all 1{,}505 test frames.

\subsection{Training and Inference}
\label{sec:supp_optimization}

The video expert is initialized from Wan2.2-5B and paired with a separately parameterized action expert in a Mixture-of-Transformers architecture~\cite{wan2025,liang2025mixtureoftransformers}. Both experts contain 30 Transformer blocks and use 24 attention heads with a head dimension of 128, giving a shared 3{,}072-dimensional attention space. Their hidden-state dimensions are 3{,}072 and 1{,}024, and their feed-forward dimensions are 14{,}336 and 4{,}096, respectively. The action expert projects its 1{,}024-dimensional hidden states into the shared attention space for queries, keys, and values, then projects the attention output back to 1{,}024 dimensions. The video expert uses a \(1\times2\times2\) spatiotemporal patch size.

\paragraph{Optimization and software.}
Additional NAVSIM training details include a per-GPU batch size of 8, AdamW \((\beta_1,\beta_2)=(0.9,0.95)\), 5\% linear warmup followed by cosine decay to 1\% of the initial learning rate, and equal coefficients of 1 for the averaged four-stream video loss and the action loss. We select checkpoints using the NAVSIM validation \texttt{pdm\_score}. Rank \(r\) seeds Python, NumPy, PyTorch, and CUDA with \(42+r\); deterministic worker seeds derive from the process seed, worker index, and rank. The software environment specifies Python 3.10 or later, PyTorch 2.7.1 with CUDA 12.8, torchvision 0.22.1 with CUDA 12.8, Accelerate 1.12.0, DeepSpeed 0.18.5, Transformers 4.49.0, Hydra 1.3.2, and NumPy 1.26.4.

\paragraph{Trajectory coordinates.}
For NAVSIM, the action target contains eight ego-frame waypoints \((x,y,\psi)\) at 0.5-s intervals. For a future global position \(\mathbf{p}_{t+h}\) and heading \(\psi_{t+h}\), we compute
\begin{equation}
\begin{aligned}
\begin{bmatrix}x_{t+h}\\y_{t+h}\end{bmatrix}
&=R(-\psi_t)\left(\mathbf{p}_{t+h}-\mathbf{p}_t\right),\\
\psi^{\mathrm{rel}}_{t+h}
&=\operatorname{wrap}_{[-\pi,\pi)}\left(\psi_{t+h}-\psi_t\right).
\end{aligned}
\label{eq:supp_trajectory_coordinates}
\end{equation}
Planar coordinates are measured in meters from the rear axle at the final observation frame. Heading is measured in radians relative to the ego orientation in that frame. The transform yields the NAVSIM ego-frame trajectory \(\mathbf a\). In the main-paper notation, \(\widetilde{\mathbf a}=\mathcal N_a(\mathbf a)=\mathbf a\), and \(\mathcal N_a^{-1}\) is also the identity map for this representation.

For WOD-E2E, the annotations provide 20 ego-frame positions at 0.25-s intervals. We augment them with headings to form 20 action targets \((x_i,y_i,\psi_i)\). With \((x_0,y_0)=(0,0)\) at the current ego-frame origin, we set \(\psi_i=\operatorname{atan2}(y_i-y_{i-1},x_i-x_{i-1})\). For displacements shorter than \(10^{-4}\)\,m, we retain the most recent valid heading, initialized to zero, and unwrap the resulting angle sequence.

\paragraph{Shifted flow-time sampling and weighting.}
We define the shift mapping
\begin{equation}
\phi_{\kappa}(\rho)
=\frac{\kappa\rho}{1+(\kappa-1)\rho},
\qquad \kappa=5.
\label{eq:supp_shift_map}
\end{equation}
During training, we independently sample
\begin{equation}
\begin{aligned}
\rho_{\mathrm{vid}},\rho_{\mathrm{act}}
&\stackrel{\mathrm{i.i.d.}}{\sim}\mathcal{U}(0,1),\\
\lambda_{g}&=\phi_{\kappa}(\rho_g),
\quad g\in\{\mathrm{vid},\mathrm{act}\}.
\end{aligned}
\label{eq:supp_time_sampling}
\end{equation}
The four visual streams share \(\lambda_{\mathrm{vid}}\), whereas the action group uses \(\lambda_{\mathrm{act}}\). The implementation passes \(N\lambda_g\) to the timestep embedding, with \(N=1000\). The equations use normalized time \(\lambda_g\in[0,1]\). We define
\begin{equation}
\begin{aligned}
q(\lambda)
&=\exp\!\left[-2\left(\lambda-\tfrac{1}{2}\right)^2\right]
-\exp\!\left(-\tfrac{1}{2}\right),\\
Z&=\frac{1}{N}\sum_{n=0}^{N-1}
q\!\left(\phi_{\kappa}\!\left(1-\frac{n}{N}\right)\right).
\end{aligned}
\label{eq:supp_weight_norm}
\end{equation}
The scheduler weight is
\begin{equation}
w(\lambda)=\frac{q(\lambda)}{Z+10^{-10}},
\label{eq:supp_scheduler_weight}
\end{equation}
which has approximately unit mean over the shifted grid.

\paragraph{Shifted Euler inference.}
For \(S\) solver steps, we use the common video-action time grid
\begin{equation}
\tau_s=\phi_{\kappa}\!\left(1-\frac{s}{S}\right),
\qquad s=0,\ldots,S,
\label{eq:supp_euler_grid}
\end{equation}
which satisfies \(\tau_0=1\) and \(\tau_S=0\). Each prediction group starts from Gaussian noise. We update each \(r\in\mathcal{R}=\{\mathrm{rgb},\mathrm{seg},\mathrm{depth},\mathrm{track},\mathrm{act}\}\) by explicit Euler:
\begin{equation}
\mathbf{y}_{\tau_{s+1}}^{r}
=\mathbf{y}_{\tau_s}^{r}
+(\tau_{s+1}-\tau_s)\widehat{\mathbf{u}}_{\theta,s}^{r},
\qquad s=0,\ldots,S-1,
\label{eq:supp_euler_update}
\end{equation}
where \(\widehat{\mathbf{u}}_{\theta,s}^{r}\) is evaluated from the current joint state with timestep embedding \(N\tau_s\).

\FloatBarrier

\end{document}